\documentclass[journal]{IEEEtran}
\ifCLASSINFOpdf
\else
\fi
\usepackage{algorithmic}
\usepackage{times}
\usepackage{epsfig}
\usepackage{graphicx}
\usepackage{amsmath}
\usepackage{amsfonts,amssymb} %hollow character

\usepackage{threeparttable}
\usepackage{stmaryrd}
\usepackage{multirow}
\usepackage{booktabs}
\usepackage{bm}

\usepackage{algorithm}

\usepackage{array}
\newcommand{\PreserveBackslash}[1]{\let\temp=\\#1\let\\=\temp}
\newcolumntype{C}[1]{>{\PreserveBackslash\centering}p{#1}}
\newcolumntype{R}[1]{>{\PreserveBackslash\raggedleft}p{#1}}
\newcolumntype{L}[1]{>{\PreserveBackslash\raggedright}p{#1}}

\DeclareGraphicsExtensions{.pdf,.jpg,.png}

\begin{document}
\title{Discriminative feature encoding for \\intrinsic image decomposition}

\author{Zongji Wang$^1$, Yunfei Liu$^2$, and Feng Lu$^{*~2,3}$% <-this % stops a space
\thanks{Correspondence should be addressed to Feng Lu.}
\thanks{$1\quad $ Key Laboratory of Network Information System Technology (NIST), Aerospace Information Research Institute, Chinese Academy of Sciences, Beijing 100190, China. wangzongji@aircas.ac.cn
   }% <-this % stops a space
\thanks{ $2\quad $ State Key Laboratory of Virtual Reality Technology and Systems, School of Computer Science and Engineering, Beihang University, Beijing 100191, China. \{lyunfei,lufeng\}@buaa.edu.cn
    }% <-this % stops a space
\thanks{ $3\quad $ Peng Cheng Laboratory, Shenzhen 518000, China.}}


\markboth{Journal of \LaTeX\ Class Files,~Vol.~14, No.~8, August~2022}%
{Shell \MakeLowercase{\textit{et al.}}: Bare Demo of IEEEtran.cls for IEEE Journals}



\maketitle

\begin{abstract}
        Intrinsic image decomposition is an important and long-standing computer vision problem. Given an input image, recovering the physical scene properties is ill-posed. Several physically motivated priors have been used to restrict the solution space of the optimization problem for intrinsic image decomposition. This work takes advantage of deep learning, and shows that it can solve this challenging computer vision problem with high efficiency. The focus lies in the feature encoding phase to extract discriminative features for different intrinsic layers from an input image. To achieve this goal, we explore the distinctive characteristics of different intrinsic components in the high dimensional feature embedding space. We define \emph{feature distribution divergence} to efficiently separate the feature vectors of different intrinsic components. The feature distributions are also constrained to fit the real ones through a \emph{feature distribution consistency}. In addition, a data refinement approach is provided to remove  data inconsistency from the Sintel dataset, making it more suitable for intrinsic image decomposition. Our method is also extended to intrinsic video decomposition based on pixel-wise correspondences between adjacent frames. Experimental results indicate that our proposed network structure can outperform the existing state-of-the-art.
\end{abstract}

\begin{IEEEkeywords}
intrinsic image decomposition; deep learning; feature distribution; data refinement.
\end{IEEEkeywords}

% \fi
\IEEEpeerreviewmaketitle

\section{Introduction}\label{sec:introduction}

In terms of intrinsic image decomposition, the albedo image $A$ indicates the surface material's reflectivity which is unchanging under different illumination conditions, while the shading image $S$ accounts for  illumination effects due to object geometry and camera viewpoint \cite{BLG2018-cnnretinex}. It is an ill-posed problem to reconstruct these two intrinsic images from a single color image $I$, which has the formation model:
\begin{equation}
    \centering
    I = A\cdot S.
    \label{intrinsic}
\end{equation}

%%% traditional solutions and principle defects
To solve this challenging inverse image formation problem, many researchers have tried  applying physically motivated priors as constraints to disambiguate the decomposition \cite{land1971-retinex,rother2011-sparsity,shen2011-sparsity,shen2008-nonlocal_texture,zhao2012-nonlocal_texture,barron2014-shape,bousseau2009-user,shen2013-user}. These methods usually represent the priors in the form of energy terms and solve the decomposition problem through graph-based inference algorithms.
With the surge of ground-truth intrinsic decomposition data \cite{grosse2009-dataset,Butler2012-dataset,bell2014-dataset}, data-driven deep learning methods \cite{BLG2018-cnnretinex,narihira2015-relative,NMY2015-DI,shi2017-nonlambertian,fan2018-revisiting,li2018-cgintrinsics} have achieved promising decomposition results and have  drawn more and more research interest. However, fully-supervised methods require high-quality and densely-labelled decompositions, which are expensive to acquire. To overcome this problem, methods training across different datasets \cite{li2018-cgintrinsics}, training on synthetic datasets \cite{shi2017-nonlambertian, li2018-cgintrinsics}, adding additional constraints \cite{fan2018-revisiting} and reusing physically motivated priors \cite{BLG2018-cnnretinex} have been proposed.

%%% our method
When developing their specific deep learning techniques,
previous methods usually extract features via a shared encoder, and then use different decoders to disentangle information for specific intrinsic layers.
Observing the different distributions between albedo and shading in the gradient domain \cite{land1971-retinex}, it is natural to assume that features representing different intrinsic layers can be separated in the embedding space. With the features separated during the encoding phase, decoders can be relieved from distilling clues for specific targets and focus on the reconstruction procedure.
This idea motivates the research in this paper.

We propose a novel two-stream encoder-decoder network for intrinsic image decomposition.
In particular, our \emph{feature distribution divergence} (FDD) constraint is designed to encourage the two encoders to extract distinctive features for different intrinsic layers. Our \emph{feature distribution consistency} (FDC) constraint is used to encourage the features of a reconstructed intrinsic layer to have a similar distribution pattern to ground-truth decompositions.
Moreover, we provide an approach to deal with the illumination inconsistency between the ground truth shading and input images in the MPI Sintel dataset, making it more suitable for intrinsic image decomposition.
We also provide an intrinsic decomposition method for video data based on  pixel-wise correspondences between adjacent frames.
This work is an extension of our previous published workshop paper \cite{wang2019single}, giving more detailed method descriptions, novel technical contributions, and more comprehensive experiments.

The major contributions of this work are:

$\bullet$ A novel two-stream encoder-decoder network for intrinsic image decomposition, in which discriminative feature encoding is achieved via  feature distribution divergence  and  feature distribution consistency constraints.

$\bullet$ A data refinement algorithm for the MPI Sintel dataset, producing a more physically consistent dataset that better suits the intrinsic decomposition task.

$\bullet$ Experimental results on various datasets to demonstrate the effectiveness of our proposed method, including experimental extension to decomposition of video data.

\section{Related work} \label{sec:Relatedwork}
\subsection{Approaches}

    Intrinsic image decomposition is a long standing computer vision problem. However, it is a seriously ill-posed problem to recover an albedo layer and a shading layer from a single color image \cite{shi2017-nonlambertian}.
    In  recent decades, considerable effort has been devoted to this challenging problem. These approaches can be coarsely classified into optimization-based methods using physically motivated priors, and deep learning based, data-driven methods \cite{shi2017-nonlambertian,bonneel2017-editing}. There are also approaches using multiple images as input \cite{weiss2001-sequences,matsushita2004-sequences,laffont2015-sequences,li2018-watching,lettry2018varying,gong2018lowrank}, treating the reflectance as a constant factor with changing illumination. These methods require the images to be captured by a static camera with varying illumination. A generative adversarial network (GAN) based domain transfer framework has also applied to image layer separation tasks \cite{liu2020separate}. Additional cues including depth maps \cite{barron2013-rgbd, chen2013-depth, lee2012-depth, kim2016-unified_depth} and near-infrared images \cite{cheng2019non} are also taken into account in some work.
    Here, we focus on key works recovering intrinsic images from a single input.
    %we would not attempt to enumerate them all but focus on the works using a single RGB image as input
    
    \subsection{Physically motivated priors based methods}
    % retinex (start point), other derived priors: sparse, low-rank,
    To solve this ill-posed intrinsic decomposition problem, researchers have derived several physically-inspired priors to constrain the solution space \cite{shi2017-nonlambertian}. Land \emph{et al.} \cite{land1971-retinex} proposed the retinex algorithm, exploring the different properties of intrinsic components in the gradient domain: large derivatives are perceived as changes in reflectance properties, while smoother variations are seen as changes in illumination.
    Based on this assumption, many priors for intrinsic image decomposition have been explored.
    Derived from a piece-wise constant property, reflectance sparsity \cite{rother2011-sparsity,shen2011-sparsity} and low-rank reflectances \cite{bousseau2009-user} have been used as constraints.
    Other constrains include the distribution difference in the gradient domain \cite{sai2015-L1Intrinsic,li2014-smooth,sheng2018step}, non-local textures \cite{shen2008-nonlocal_texture,zhao2012-nonlocal_texture}, shape and illumination \cite{barron2014-shape}, and user strokes \cite{bousseau2009-user,shen2013-user}.
    These hand-crafted priors are not likely to be valid across complex datasets \cite{bonneel2017-editing}.
    Bi \emph{et al.} \cite{sai2015-L1Intrinsic} presented an approach using the $L_1$ norm for piece-wise image flattening, and proposed an algorithm for complex scene-level intrinsic image decomposition. Li \emph{et al.} \cite{li2014-smooth} presented a method to automatically extract two layers from an image based on differences in their distributions  in the gradient domain. Sheng \emph{et al.} \cite{sheng2018step} proposed an approach based on illumination decomposition, in which a shading image is decomposed into drift and step shading channels based on different distribution properties in gradient domain. Based on an analysis of the logarithmic transformation, Fu \emph{et al.} \cite{fu2016srie} introduced a weighted variational model to refine the regularization terms for intrinsic image decomposition. Later, Fu \emph{et al.} \cite{fu2019towards} presented an algorithm incorporating a reflectance sparseness regularizer based on the $L_0$ norm and a shading smoothness regularizer based on total variation.
    Non-local texture constraints \cite{shen2008-nonlocal_texture,zhao2012-nonlocal_texture} are used to find pixels with the most similar reflectance within an image.
    Krebs \emph{et al.} \cite{krebs2020intrinsic} developed a method for intrinsic image decomposition from a single RGB or multispectral image, taking the mathematical properties of the mean and standard deviation along the spectral axis into consideration.
    Although these methods restrict the solution space to a feasible region, such specifically designed priors cannot hold under complex conditions. The results are largely influenced by the parameter settings, which need expert knowledge. Our method differs in that it is data driven, and parameters are automatically learned from  the dataset.
   
   \begin{figure*}[!t]
    \centering
    \includegraphics[width=1.0\linewidth]{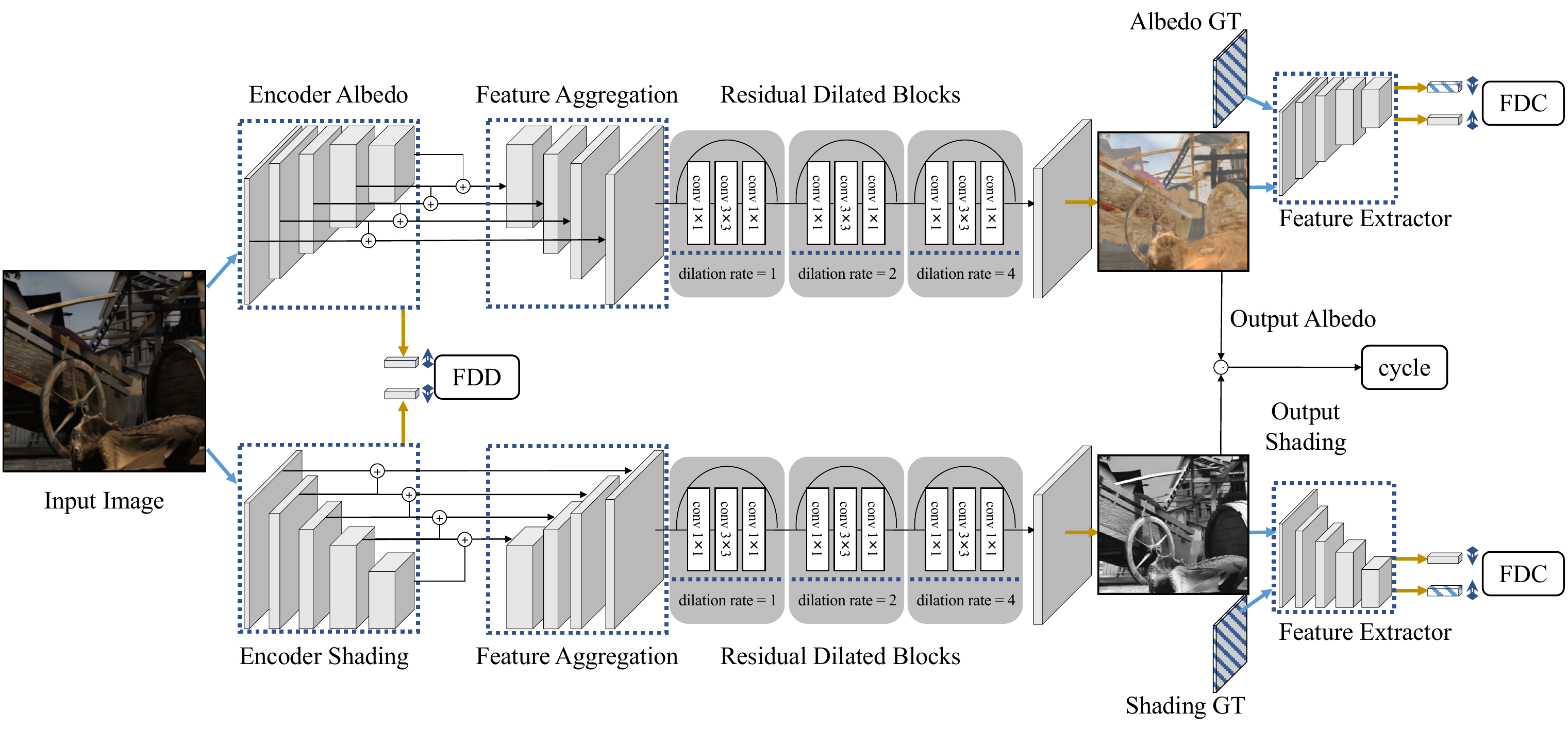}
    \caption{Framework of our two-stream intrinsic image decomposition network. The input image is passed through two sub-network streams for albedo and shading image reconstruction respectively. We use the extractor in VGG-19 as the encoder structure, which extracts multi-scale feature maps. These are then aggregated by sequences of upsampling, concatenation, and convolution. Finally, three residual dilated blocks are used as a decoder to reconstruct intrinsic images from the fused feature maps. $\oplus$ denotes feature aggregation, $\odot$ denotes element-wise multiplication, and rounded boxes represent loss computations. \emph{Cycle, FDC} and \emph{FDD} mean cycle loss, feature distribution consistency and feature distribution divergence respectively.}
    \label{fig:1}
\end{figure*}

    \subsection{Deep learning methods}
    Thanks to the public availability of intrinsic image datasets such as the MIT intrinsic dataset \cite{grosse2009-dataset}, the MPI Sintel dataset \cite{Butler2012-dataset} and Intrinsic Images in the Wild (IIW) \cite{bell2014-dataset}, application of deep learning to intrinsic decomposition has surged \cite{narihira2015-relative, Tang2012-DLN,  zhou2015-datadriven, zoran2015-ordinal, nestmeyer2017-filtering}. Direct intrinsics \cite{NMY2015-DI} provided the first entirely deep learning model that directly outputs  albedo and shading layers given a color image. Results from this method are blurred due to down-sampling during encoding and  deconvolution during decoding.
    
    Facing the fact that high-quality and densely-labelled intrinsic images are expensive to acquire, many methods have been developed to train models with additional constraints \cite{fan2018-revisiting, fu2021multi}, reusing physically motivated priors \cite{BLG2018-cnnretinex, seo2021deep, baslamisli2021physics, baslamisli2021shadingnet,zhu2021derendernet}, expanding the dataset with synthetic images \cite{shi2017-nonlambertian,sial2020deep,li2018-cgintrinsics} or training across datasets \cite{li2018-cgintrinsics}.
    
    Fan \emph{et al.} \cite{fan2018-revisiting} provided a network structure using a domain filter between  edges in the guidance map to encourage   piece-wise constant reflectance. \cite{fu2021multi} used region masks to guide the separation of different intrinsic components. By utilizing additional constraints, the solution space for the problem is further restricted.
    Seo \emph{et al.} \cite{seo2021deep} proposed an image-decomposition network, which makes use of all the three premises regarding consistency from retinex theory. In this work, pseudo images (generated color-transferred multi-exposure images) are used for training.
    Baslamisli \emph{et al.} \cite{BLG2018-cnnretinex} presented a two-stage framework to firstly split the image gradients into albedo and shading components, which are then fed into decoders to predict pixel-wise intrinsic values. In their later work \cite{baslamisli2021physics}, the gradient descriptors for albedo and shading are derived from a physics-based reflection model and used to compute the shading map directly from  RGB image gradients.
    \cite{baslamisli2021shadingnet} and \cite{zhu2021derendernet} derived fine-grained shading components from a physics-based image formation model, in which the shading component is further decomposed into direct and indirect components, and shape-dependent/independent ones.
    These works proposed novel methods by revisiting physically motivated priors.
    Shi \emph{et al.} \cite{shi2017-nonlambertian} trained a model to learn albedo, shading and specular images on a large-scale object-level synthetic dataset by rendering ShapeNet \cite{chang2015-shapenet}. 
    Sial \emph{et al.} \cite{sial2020deep} trained an intrinsic decomposition model on a synthetic ShapeNet-based scene dataset which has more realistic lighting effects.
    Li \emph{et al.} \cite{li2018-cgintrinsics} presented an end-to-end learning approach that learns better intrinsic image decomposition by leveraging datasets with different types of labels.
    
    The majority of these methods extract features via a shared encoder, and then use different decoders to disentangle information for specific intrinsic layers.
    In contrast to these works, we try to exploit the difference between intrinsic components in feature space through a novel two-stream framework. With the features separated in the encoding phase, decoders can be relieved from distilling clues for specific targets and focus on the reconstruction procedure.

    \subsection{Intrinsic video}
    Kong \emph{et al.} \cite{kong2014intrinsic} defined intrinsic video estimation as the problem of extracting temporally coherent albedo and shading from video alone.
    Ye \emph{et al.} \cite{ye2014intrinsic} proposed a probabilistic approach to propagate the reflectance from the initial intrinsic decomposition of the first frame.
    In order to achieve temporal consistency, these methods rely on optical flow to provide  correspondences across time.
    Meka \emph{et al.} \cite{meka2016live} presented the first approach to tackle the hard intrinsic video decomposition problem at real-time frame rates. This method applies global consistency constraints in  space and time  based on random sampling.
    Lei \emph{et al.} \cite{lei2020blind} presented a novel and general approach for blind video temporal consistency. This method is only trained on a pair of original and processed videos directly instead of a large dataset.
    In this paper, we simply extend our intrinsic image decomposition method to video based on optical flow, preserving temporal consistency during the decomposition process.

\section{Method}\label{sec:approach}

\begin{figure}[!t]
    \centering
    \includegraphics[width=\linewidth]{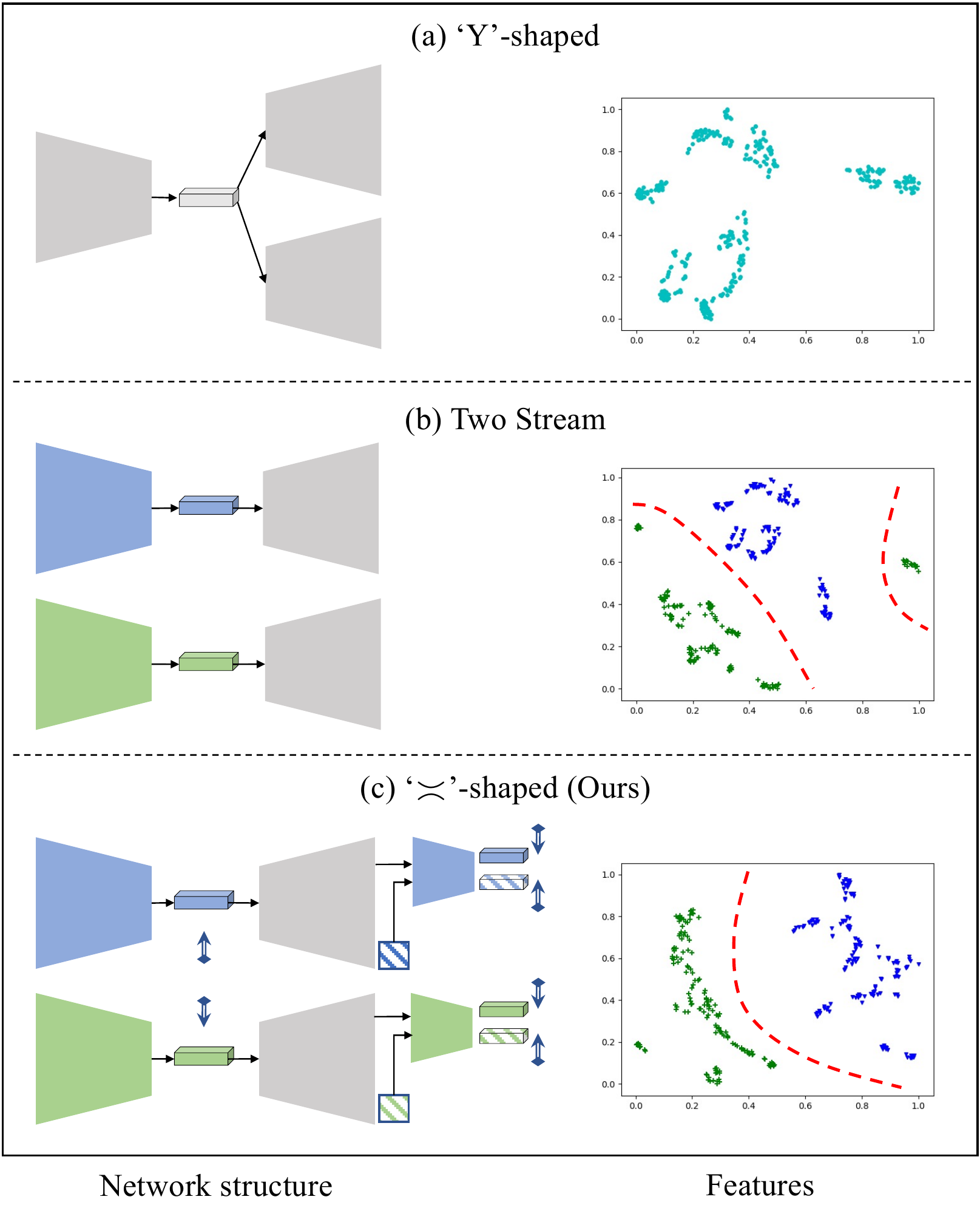}    \caption{Feature distributions of different network structures. In each row,  left: simplified network structure,  right: features visualized by t-SNE. The dotted curve represents the separating boundary of different domains in  feature space. (a): `Y'-shaped framework using a shared encoder. (b): plain two stream framework using two independent encoder-decoder sub-networks. (c): our `$\asymp$'-shaped framework, in which  features from different streams interact during training.}
    \label{fig:2}
\end{figure}

%%% Network structure
\subsection{Network structure}
Our network architecture is visualized in Figure \ref{fig:1}. The framework consists of two streams of encoder-decoder sub-networks. One performs albedo image reconstruction, and the other, shading image reconstruction. Taking the albedo sub-network for example, the input image is passed through a convolutional encoder to extract multi-level features, which are then aggregated by sequences of upsampling, concatenation, and convolution. In the decoding phase, the fused multi-scale features are fed into a sequence of three residual dilated blocks to reconstruct the albedo intrinsic image. The shading sub-network has the same structure as the albedo sub-network. In practice, we adopt VGG-19 \cite{Simonyan2014-vgg} pretrained on ImageNet \cite{deng2009-imagenet} as the initial encoder.

Previous works usually use a shared encoder to extract features containing both albedo and shading information. Different decoders are then applied to distill clues from the comprehensive features for specific intrinsic image reconstruction. The `Y'-shaped framework can be formulated as:
\begin{equation}
    \begin{aligned}
        A &= g(f(I;~\Theta); ~\Omega_a) = g_a \circ f(I), \\
        S &= g(f(I;~\Theta); ~\Omega_s) = g_s \circ f(I),
    \end{aligned}
    \label{shared_encoder}
\end{equation}
where $f(\cdot;~\Theta)$ and $g(\cdot;~\Omega)$ denote the feature encoder and decoder respectively. $\Theta$ and $\Omega$ represent their corresponding trainable parameters.
In this equation, we intend to represent $g(\cdot, \Theta_a)$ as $g_a(\cdot)$, in which the subscript of the function symbol $f_*$ means that the trainable parameters $\Theta_*$ are different with respect to specific intrinsic components, while the network frameworks are the same.

Unlike such methods, our network design has two encoders for albedo and shading images respectively. In this paper, we denote this structure as an `$\asymp$'-shaped framework:
\begin{equation}
    \begin{aligned}
        A &= g(f(I;~\Theta_a); ~\Omega_a) = g_a \circ f_a(I), \\
        S &= g(f(I;~\Theta_s); ~\Omega_s) = g_s \circ f_s(I).
    \end{aligned}
    \label{differ_encoder}
\end{equation}
Using this framework, the encoders ($f_a(\cdot)$, $f_s(\cdot)$) are able to extract features more pertinent to their reconstruction targets (albedo, shading). In Figure \ref{fig:2}, we visualize the feature distributions of different network structures, which explains our idea  pictorially.

    In each row of Figure \ref{fig:2}, a feature embedding visualization using t-SNE is provided on the right. Each data point represents a feature extracted by the encoder, which is then fed into the corresponding intrinsic image reconstruction decoder. We use the same color coding for the embedding's data points  and the extracted feature vectors from the simplified network structure. The proposed `$\asymp$'-shaped framework (given in detail in Figure \ref{fig:1}) results in a better feature embedding. For instance, in the embedding space, the features for different intrinsic components are better separated.

The rest of this section introduces the core idea and detailed design  of the discriminative feature encoding. Then, important constraints for our intrinsic decomposition network are explained.

\subsection{Discriminative feature encoding}
\subsubsection{Basis}
%Gradient space discriminant. [Deep retinex] --> Feature space
Our work is inspired by Land et al. \cite{land1971-retinex}: the retinex approach assumes that albedo and shading layers possess different properties in the gradient domain. By utilizing such discriminative properties, the intrinsic decomposition results can be improved.
In this work, we study and exploit the discriminative properties in a more general convolutional feature space.
We next describe the proposed discriminative feature encoding in detail.

\subsubsection{Feature distribution divergence}
As Figure \ref{fig:1} shows, the encoding phase consists of multiple (convolution, relu, maxpooling) blocks, through which the input signal is encoded into several different abstraction levels. The multi-scale features are denoted  $\{\bm{f^{E_1}},\dots,\bm{f^{E_n}}\}$, in which $\bm{f^{E_i}}$ represents the output feature of the $i^{th}$ block. We define the feature distance function as $d: \mathbb{R}^{m\times n \times c} \times \mathbb{R}^{m \times n \times c}\mapsto \mathbb{R}$, where $c$ denotes the feature channel number and  the input signal has spatial size $m\times n$:
\begin{equation}
    \begin{aligned}
        &d_{\cos}(\bm{f_{a}^{E_i}}, \bm{f_{s}^{E_i}}) = \\
            &\qquad\frac{1}{N_i} \sum_{\forall (x,y)} \left(\frac{<\bm{f_{a}^{E_i}}(x,y), \bm{f_{s}^{E_i}}(x,y)>}{||\bm{f_{a}^{E_i}}(x,y)||_2\,||\bm{f_{s}^{E_i}}(x,y)||_2} \right)^2, \\
        &d_{L_1}(\bm{f_{a}^{E_i}}, \bm{f_{s}^{E_i}}) = h(||\bm{f_{a}^{E_i}} - \bm{f_{s}^{E_i}}||_1), \\
        &d(\bm{f_{a}^{E_i}}, \bm{f_{s}^{E_i}}) = \alpha  d_\mathrm{cos}(\bm{f_{a}^{E_i}}, \bm{f_{s}^{E_i}}) + \beta d_{L_1}(\bm{f_{a}^{E_i}}, \bm{f_{s}^{E_i}}).
    \end{aligned}
    \label{feature_distance}
\end{equation}
Feature distance measurement is based on the cosine distance between two vectors and $L_1$ norm.
In Eq. (\ref{feature_distance}), $\bm{f_{a}}$ and $\bm{f_{s}}$ represent features from the albedo encoder and the shading encoder respectively.
$<\cdot, \cdot>$ is the inner product in Euclidean space, $N_i = m_i\times n_i$;, and $(x,y)$ represents a spatial location in a feature map. $h(\cdot)$ is a distance rescaling function  in the form of a modified  sigmoid function to ensure $d_{L_1}\in (0,1)$. We use 
\[
h(d)=1-\frac{1}{1+\exp{(-(d-1.2 \exp(1.2))/1.2^2)}}.
\]

The feature distribution divergence $\mathcal{L}_\mathrm{fdd}$ is now formulated as:
\begin{equation}
    \begin{aligned}
        \mathcal{L}_\mathrm{fdd} = \sum_{i}^{n} \omega_i  d(\bm{f_{a}^{E_i}}, ~\bm{f_{s}^{E_i}}),
    \end{aligned}
    \label{feature_divergence_loss}
\end{equation}
where $\omega_i \geq 0$ is the weight for the feature distance for abstraction level $i$. Empirically, we use five different levels of abstraction in our experiments ($n=5$). We set $[\omega_1,\dots,\omega_5]=[0.01, 0.1, 0.5, 0.7, 1.0]$ and $\alpha=0.3$, $\beta=0.1$.

\subsubsection{Feature distribution consistency}
The feature distribution divergence aims to increase the distance between the feature vectors embedded by different encoders. However, this is not sufficient for discriminative feature encoding.
The core idea of Fisher's linear discriminant is to maximize the distance between classes and minimize the distance within classes simultaneously.
As an analogue of that, along with the feature distribution divergence described above, we use the feature perceptual loss \cite{johnson2016-perceptual} between the predicted and ground truth intrinsic images to constrain the encoding process, encouraging the embedded features to fit the real distribution.

We use the same distance measurement as for the feature distribution divergence. $d(\bm{f_\mathrm{pred}^{E_i}}, \bm{f_\mathrm{real}^{E_i}})$ denotes the feature distance in the $i_{th}$ abstraction level.

The feature distribution consistency $\mathcal{L}_\mathrm{fdc}$ is formulated as:
    \begin{equation}
        \begin{aligned}
            \mathcal{L}_\mathrm{fdc} = \sum_{i}^{n} \gamma_i \left(\left(1-d(\bm{f_{pred,a}^{E_i}}, \bm{f_{real,a}^{E_i}})\right)\right. + \\ \left.\left(1-d(\bm{f_{pred,s}^{E_i}}, \bm{f_{real,s}^{E_i}})\right)\right),
        \end{aligned}
        \label{feature_distribution_constraint}
    \end{equation}
where $\gamma_i \geq 0$ is a weight. Note that $(1-d(\bm{f_\mathrm{pred}}, \bm{f_\mathrm{real}})) \in (0,1)$ represents the feature similarity between $\bm{f_\mathrm{pred}}$ and $\bm{f_\mathrm{real}}$. Minimizing Eq. (\ref{feature_distribution_constraint}) encourages the predicted and ground truth intrinsic images to have similar perceptual features. In practice, the encoders are reused to extract features from the predicted and target results in our framework, so the embedded feature distribution can be optimized directly during training. Empirically, we set $[\gamma_1,\dots,\gamma_5]=[1.0, 1.0, 1.0, 1.0, 1.0]$ and $\alpha=0.1$, $\beta=0.9$.

\subsection{Basic supervision constraints}
\subsubsection{Use of losses}
Besides the above constraints for discriminative feature encoding, several basic supervision losses are adopted to train the intrinsic image decomposition network.

As described in Eq. (\ref{differ_encoder}), given an image $I$, the albedo image $A$ and the shading image $S$ are predicted through trained $g_{a}\circ f_{a}$ and $g_{s}\circ f_{s}$. With densely-labelled intrinsic images $\widehat{A}$ and $\widehat{S}$ as  ground truth data, we constrain the pixel-wise predictions using the reconstruction loss $\mathcal{L}_\mathrm{rec}$ and the gradient loss $\mathcal{L}_\mathrm{grad}$.

\subsubsection{Reconstruction loss}
We use the $L_1$ loss $\mathcal{L}_{L_1}$ combined with the SSIM (structural similarity index \cite{wang2004-ssim}) loss $\mathcal{L}_\mathrm{SSIM}$ as the reconstruction loss:
%Thus we use the SSIM loss $\mathcal{L}_\mathrm{SSIM}$ as the mid-level difference. The reconstruction loss is the combine of $L_1$ loss and SSIM loss:
    \begin{equation}
        \begin{aligned}
            \mathcal{L}_\mathrm{rec}  &= \lambda_{L_1} \mathcal{L}_{L_1} + \lambda_\mathrm{SSIM} \mathcal{L}_\mathrm{SSIM},\\
            \mathcal{L}_{L_1}  &= ||A-\widehat{A}||_1 + ||S-\widehat{S}||_1 + ||A \times S - I||_1,\\
            \mathcal{L}_\mathrm{SSIM} &= (1-\mathrm{SSIM}(A,\widehat{A})) + (1-\mathrm{SSIM}(S,\widehat{S}))\\
            &+ (1-\mathrm{SSIM}(A \times S, I)),
        \end{aligned}
        \label{reconstruction_loss}
    \end{equation}
where $\mathrm{SSIM}(\bm{x}, \bm{y})$ measures the structural similarity of images $\bm{x}$ and $\bm{y}$. The SSIM loss is $(1-\mathrm{SSIM}(\bm{x},\bm{y}))$, indicating structural dissimilarity. Empirically, we set $\lambda_{L_1}=30.0$ and $\lambda_\mathrm{SSIM}=0.5$.  $A \times S$ is computed pixel-wise.

The cycle loss is used to encourage the product of the predicted $A$ and $S$ to be similar to that of the input image $I$.

\subsubsection{Gradient loss}
We also use  image gradients as supervision to help preserve details in the intrinsic images:
\begin{equation}
    \begin{aligned}
        \mathcal{L}_\mathrm{grad} = ||\nabla_{x}A - \nabla_{x}\widehat{A}||_2^2 + ||\nabla_{y}A - \nabla_{y}\widehat{A}||_2^2 +\\
        ||\nabla_{x}S - \nabla_{x}\widehat{S}||_2^2 + ||\nabla_{y}S - \nabla_{y}\widehat{S}||_2^2,
    \end{aligned}
    \label{gradient_loss}
\end{equation}
where $\nabla_{x}, \nabla_{y}$ are the image gradients in the  $x$ and $y$ directions.

For datasets with ground truth decomposition like the MIT intrinsic dataset and  MPI Sintel, the total loss is constructed as:
\begin{equation}
    \begin{aligned}
        \mathcal{L}_\mathrm{total} = \lambda_{1} \mathcal{L}_\mathrm{rec} + \lambda_{2} \mathcal{L}_\mathrm{grad} + \lambda_{3} \mathcal{L}_\mathrm{fdd} + \lambda_{4} \mathcal{L}_\mathrm{fdc}.
    \end{aligned}
    \label{total_loss}
\end{equation}
Empirically, we set $[\lambda_1,\dots,\lambda_4]=[1.0, 1.5, 0.1, 1.0]$.

    \subsection{Adjustment for sparsely-labelled data}
    \subsubsection{Sparse labelling}
As well as the densely-labelled datasets mentioned above, recently, the sparsely-labelled dataset IIW has become available, with a larger number of real-world images. To apply our core idea to this situation, we must slightly adjust the training framework, as we first explain, and then give the loss functions measuring the intrinsic image reconstruction quality.
    
    \begin{figure*}[!t]
        \centering
        \includegraphics[width=\linewidth]{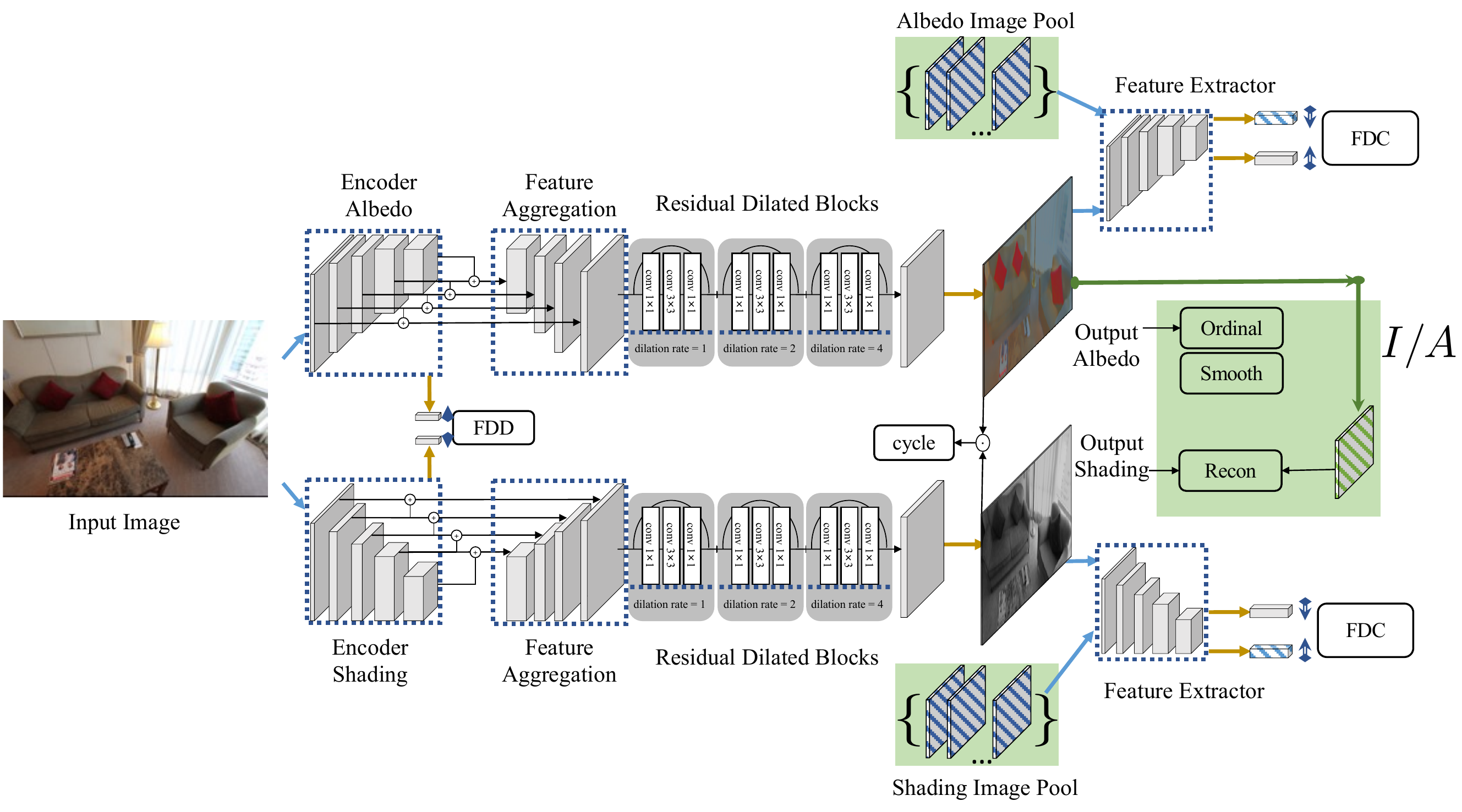}         \caption{Framework adjustment for sparsely-labelled data. Changed parts are highlighted in green.}
        \label{fig:3}
    \end{figure*}
    
    \subsubsection{Framework adjustment}
    In order to train on the sparsely-labelled IIW dataset, there are a few barriers to overcome. Figure \ref{fig:3} shows the adjustments made to our framework.
    
    One problem is the lack of dense supervision during the albedo reconstruction procedure. The reconstruction loss in Eq. (\ref{reconstruction_loss}) and the gradient loss in Eq. (\ref{gradient_loss}) both need pixel-wise dense supervision. However, the IIW dataset only has sparse annotations of reflectance comparisons at selected points of the images. Therefore, we have to apply alternative constraints utilizing sparse labelling. Specifically, we use the ordinal loss to measure the difference between the output albedo image and the annotated input. Furthermore, smoothness constraints are applied to model the smoothness prior of the albedo component.
    
    A further problem is the lack of shading ground truth, which is necessary in our two-stream training framework. The core idea of the feature distribution divergence is to extract distinctive features corresponding to different target intrinsic layers from the same input image. However, there is no annotation for the shading layers in this dataset. To circumvent the lack of ground truth data, we directly synthesize the target shading image from the input $I$ and the reconstructed albedo $A$, using Eq. (\ref{intrinsic}). Therefore, the synthesized shading image can be used as dense supervision.
    
    Last but not least, the lack of reference intrinsic images causes problems for computing the feature distribution consistency. The FDC is designed to constrain the intrinsic image to have a similar distribution to the corresponding real reference. In detail, this constraint is achieved by minimizing the feature perceptual loss between the reconstructed and reference images. However, dense ground truth images are not provided in the IIW dataset, making ground truth features unavailable. To solve this problem, we maintain an image pool for each training stream, in which batches of reconstructed intrinsic images are gathered as the guidance for feature perceptual loss.
    
    \subsubsection{Training constraints}
    As noted, we thus modify the constraints to suit the sparse annotations in the IIW dataset. We now describe the ordinal loss and the smoothness constraints used.
    % to measure the difference between the predicted and the target intrinsic images.
    
We first consider the ordinal loss.
    Since dense ground truth labels are not available in \cite{bell2014-dataset}, it introduced the \emph{weighted human disagreement rate} (WHDR) as an error metric. Following \cite{li2018-cgintrinsics}, we use an ordinal loss based on WHDR as the sparse supervision term. 
    
    The ordinal loss $\mathcal{L}_\mathrm{ord}$ is obtained by accumulating all the annotated pairs in the albedo image:
    \begin{equation}
        \begin{aligned}
            \mathcal{L}_\mathrm{ord} = \sum_{(i,j)}e_{i,j}(A),
        \end{aligned}
        \label{ordinal_loss}
    \end{equation}
where $e_{i,j}(A)$ represents the error for a pair of annotated pixels $(i,j)$ in the predicted albedo image $A$. A detailed definition is provided in Appendix A.
    
We next consider smoothness priors: we adopt the same ones as \cite{li2018-cgintrinsics}.
    The smoothness constraints comprise the albedo smoothness constraint $\mathcal{L}_\mathrm{asmooth}$ and the shading smoothness constraint $\mathcal{L}_\mathrm{ssmooth}$. The albedo component is constrained using a multi-scale $L_1$ smoothness term, through which the albedo layer reconstruction is encouraged to be piecewise constant. Shading smoothness is constrained using a densely-connected $L_2$ term. Detailed definitions are provided in Appendix A.
    
The total loss for sparsely-labelled data is:
    \begin{equation}
        \begin{aligned}
            \mathcal{L}_\mathrm{total}^\mathrm{sparse} = &\mathcal{L}_\mathrm{total}^{\backslash\hat{A}} + \lambda_\mathrm{ord}\mathcal{L}_\mathrm{ord} + \\
            &\lambda_\mathrm{As} \mathcal{L}_\mathrm{asmooth} + \lambda_\mathrm{Ss} \mathcal{L}_\mathrm{ssmooth},
        \end{aligned}
        \label{total_loss_sparse}
    \end{equation}
    where $\mathcal{L}_\mathrm{total}^{\backslash\hat{A}}$ means the total loss function $\mathcal{L}_\mathrm{total}$ for densely-labelled data in Eq. (\ref{total_loss}), removing the terms containing dense albedo ground truth $\hat{A}$. $\lambda_\mathrm{As}$ and $\lambda_\mathrm{Ss}$  weight  the smoothness priors. A study considering different values of $\lambda_\mathrm{As}$ and $\lambda_\mathrm{Ss}$ is provided in \ref{sec:iiw_experiments}.

    \subsection{Adjustment for video data}
    The MPI Sintel dataset is composed of short films, so it naturally has temporal consistency. We also investigated a suitable framework for intrinsic decomposition of such video data.
    
    For video intrinsic decomposition, adjacent pairs of frames are input into our two-stream networks to get the corresponding intrinsic layers. In addition to using the single image decomposition framework, optical flow is computed from the sequential input images, and used to provide temporal consistency guidance for the output intrinsic layers.
     
    Optical flow is typically used to construct temporal correspondences between two adjacent frames, assuming that the pixel intensities of an object do not change between consecutive frames, and that neighbouring pixels have similar motion. In this work, the optical flow field of pairs of consecutive input images is obtained directly from the MPI Sintel dataset. Besides the single image intrinsic decomposition loss $\mathcal{L}_\mathrm{total}$ in Eq. (\ref{total_loss}), the optical flow $\bm{u}$ is used to enhance the temporal consistency of the output intrinsic layers:
    \begin{equation}
        \begin{aligned}
            \mathcal{L}_\mathrm{temporal} = &\lambda_A \sum_i \omega_{\bm{u}}(i)||A^{t+1}(i+\bm{u}(i)) - A^{t}(i)||_1 +\\
            &\lambda_S \sum_i \omega_{\bm{u}}(i)||S^{t+1}(i+\bm{u}(i)) - S^{t}(i)||_1,
        \end{aligned}
        \label{loss_temporal}
    \end{equation}
    where $A^{t}(i)$ is the $i^{th}$ pixel in the albedo image of frame $t$. We denote the optical flow map between the consecutive frames $(t, t+1)$ as $\bm{u}$. The mask $\omega_{\bm{u}}$ records which pixels have valid optical flow value. $\omega_{\bm{u}}(i)=0$ if  pixel $i$ is occluded, and is $1$ otherwise. We use $\lambda_A$ and $\lambda_S$ to balance the importance of the temporal consistency terms between albedo and shading layers.
    
    % need ablation study.

\section{Intrinsic data refinement}\label{sec:data_refine}
\subsection{Basic approach}
The MPI Sintel dataset \cite{Butler2012-dataset} is a publicly-available densely-labelled dataset containing complex indoor and outdoor scenes.
%It has been used as training data and evaluation benchmarks for intrinsic decomposition.
It was originally designed for optical flow evaluation. For research into intrinsic image decomposition, ground truth shading images have been rendered with a constant gray albedo considering illumination effects. However, due to the creation process, the original input frames can not be reconstructed from the ground truth albedo and shading layers through Eq. (\ref{intrinsic}).

As  the first row of Figure \ref{fig:5} shows later, the specular component of the shading image cannot be observed in the original image, which means it does not share the same illumination condition. Although the simplified image formation model Eq. (\ref{intrinsic}) need not be strictly respected, it is  physically incorrect to extract a shading layer depicting different illumination effects from the original image. To overcome this inconsistency, previous works \cite{NMY2015-DI} directly resynthesize original images $I$ from the ground truth albedo $A$ and shading $S$ via Eq. (\ref{intrinsic}). However, this approach does not deal with the specular component of the shading layer, which is considered not to be modeled well by Eq. (\ref{intrinsic}) \cite{shi2017-nonlambertian}.

In this paper, we propose an approach to refine the dataset in order to shift it into a domain more representative of real images. The refined MPI Sintel dataset (MPI\_RD) is subject to the image formation model in Eq. (\ref{intrinsic}), and the shading layers contain no color information (gray shading). In addition, the shading layers in the MPI\_RD maintain  consistency with the original images. This can be shown in two ways. For one thing, the specular component is removed from the shading layer. For another thing, shape details observed in the original images are preserved in the shading layer. We describe our data refinement algorithm in \ref{alg:1}.
In summary, we shift the distribution of the albedo layer to a higher mean value, and then reconstruct the shading layer from the original image and the shifted albedo (steps 1 to 5). Next, invalid pixels in the reconstructed shading layer are computed using local linear embedding (LLE) \cite{roweis2000-lle} with the input $I$ as the guiding image adopted to construct the embedding weights (steps 6 to 7). Finally, the input image is resynthesized from the processed albedo and shading images (step 8). % need to update step index

\begin{algorithm}[t!]
    \begin{algorithmic}

    \STATE \textbf{Input:}
       Original MPI Sintel dataset comprising input images $I$, albedo images $A$ and shading images $S$
    \STATE \textbf{Output:}
        Refined MPI Sintel dataset comprising $I^*,A^*,S^*$, so that $I^*=A^*  S^*$,  constrained by the intrinsic decomposition model in Eq. (\ref{intrinsic})
        
        \FOR{each $i \in [1, N]$}
        \STATE 1: convert the RGB image into $L^*a^*b^*$ space, and extract the $L$ channel as $\{I_i, A_i, S_i\}$ ;
        \STATE 2: reconstruct the albedo and shading:  $ \hat{A_i}=I_i/S_i$, $\hat{S_i}=I_i/A_i$;
        \STATE 3: compute the validity mask for $\hat{A_i}$:  $ M_i = (0<\hat{S_i}<1) ~\&~ (0<\hat{A_i}<1)$;
        \STATE 4: compute the statistics of valid pixels according to $M_i$ from  $\hat{A_i}$: $(\hat{\mu}, \hat{\sigma})$;
        \STATE 5: shift the distribution of $A_i$ ${\mu, \sigma}$ to the reconstructed valid statistics:  $ \tilde{A_i} = \hat{\sigma}(A_i - \mu)/{\sigma}  + \hat{\mu};$
        \STATE 6: reconstruct the shading from $\tilde{A_i}$: $\tilde{S_i}=I_i / \tilde{A_i};$
        \STATE 7: reconstruct the invalid pixels in $\tilde{S_i}$ with the help of $I_i$: $S_i^*$;
        \STATE 8: convert $\tilde{A_i}$ into RGB color space: ${A_i^*}$, and reconstruct $I_i^* = A_i^*  S_i^*$;
        \ENDFOR
    \end{algorithmic}
    \caption{Data refinement for MPI Sintel
    }\label{alg:1}  %        \label{refine_data}
\end{algorithm}
    
\subsection{Improvements}
    Although the data refinement algorithm \ref{alg:1} successfully suppresses inconsistencies in the intrinsic components in the MPI dataset,   problems still remain in terms of temporal consistency.
    For example, we can observe intensity jittering in consecutive albedo images, and  areas lacking detail in shading images.
   We now discuss  possible causes for these defects, as determined by   investigating the dataset, and consequently provide  methods to alleviate them. 
    
    Jittering effects can be observed in frames with notable statistical changes (mean color changes caused by large dark object occlusion).
    Areas lacking detail are usually observed in shading images whose source image has low contrast areas.
    Note that in the single-image refinement procedure, the validity mask is computed based on the source image and the original albedo and shading images. Invalid pixels (pixel values $\notin (0,1)$) are directly truncated, and therefore  not used when computing statistics of the albedo image. Such invalid areas in the shading image are then reconstructed using the LLE algorithm. In fact, the invalid areas lacking detail often overlap with the LLE reconstructed areas, resulting in low quality reconstruction.
    In conclusion,  jittering effects are mainly due to large dark region changes between frames, while lack of detail is mainly caused by invalid region reconstruction artifacts.
    
This analysis results in  a simple method to deal with those problems.
    To avoid intensity jittering between consecutive frames, we use optical flow to construct the correspondences between frames, and expand the pixel set for statistical computation. This  increases robustness of the image statistics.
    To overcome lack of detail in shading images, the region reconstruction method is optimized to take temporal correspondences into consideration.

\subsection{Temporal consistency measurement} 
To measure temporal consistency in the sequential data, we use the same video temporal consistency metric (TCM) as \cite{yao2017occlusion}:
    \begin{equation}
        \begin{aligned}
            \mathrm{TCM}_t = \exp\left(-\left|\frac{|| O_t - \mathrm{warp}(O_{t-1}) ||^2}{|| V_t - \mathrm{warp}(V_{t-1}) ||^2}-1\right|\right),
        \end{aligned}
        \label{Eq_tcm}
    \end{equation}
    where $O_t$ and $V_t$ represent the $t^{th}$ frame in the output video ($O$) and the input video ($V$) respectively.
    $\mathrm{warp}()$ is the warping function using the optical flow. The TCM of the $t^{th}$ frame is calculated using the warping error between frames. The 2-norm of a matrix $||\cdot||$ is the sum of squares of its elements. Through this equation, the processed video ($O$) is encouraged to be temporally consistent according with variations in the input video.

    In order to visualize the video processing effects, we record pixelwise temporal consistency value in a TCM map, computed as:
    \begin{equation}
        \begin{aligned}
            \mathrm{Map}_t(\cdot) = \exp\left(-\left|\frac{(O_t(\cdot) - \mathrm{warp}(O_{t-1})(\cdot))^2}{(V_t(\cdot) - \mathrm{warp}(V_{t-1})(\cdot) )^2}-1\right|\right),
        \end{aligned}
        \label{Eq_tcmmap}
    \end{equation}
    where the TCM score is computed per pixel.

\begin{figure*}[t!]
    \centering
    \includegraphics[width=\linewidth]{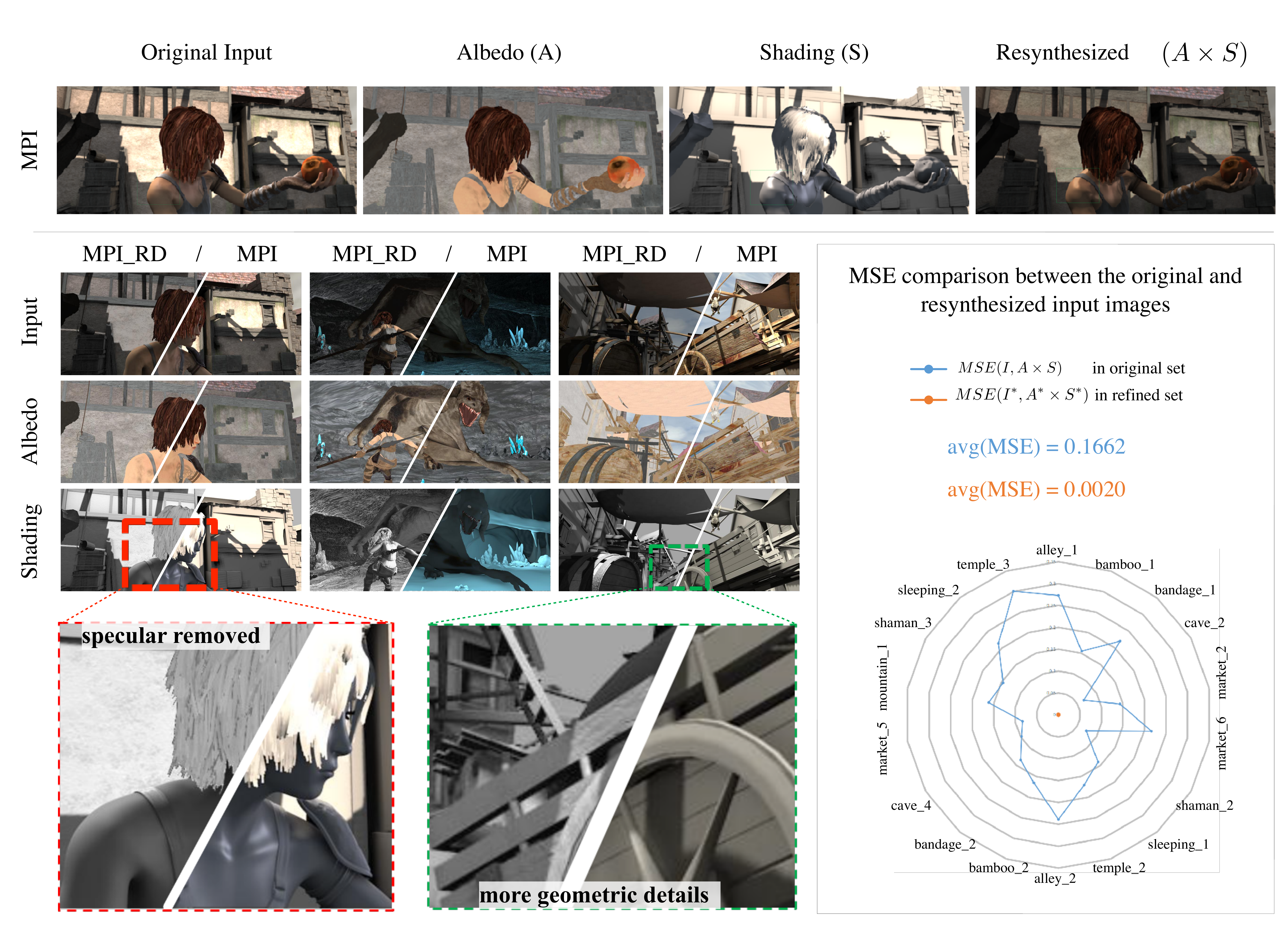}    \caption{Refined  (MPI\_RD) and  original (MPI) MPI Sintel datasets. The top row shows an example of illumination inconsistency in  MPI. At the bottom is an example for  MPI\_RD. At the bottom left, each image is split into two parts: the left shows the refined data, and the right is the original data. The close ups  show detail regions in the shading images, which in MPI\_RD exclude specular components and preserve more geometric details. At bottom right is an MSE comparison between the original and resynthesized images. The MSE value for MPI\_RD is significantly lower than for MPI.}
    \label{fig:4}
%    \label{mpi_rd_dataset}
\end{figure*}

\section{Results and discussion}\label{sec:Results}
%%% Datasets
\subsection{Datasets}
\subsubsection{MPI Sintel dataset and our refined version}
%%%%%%%%%%%%%%%%%%%%%%%%%%%%%%%%% This part
%Show some results, before and after refined, statistics (be subject to I = A*S) and visual results.
Sintel is an open source 3D animated short film, which has been published in many formats for various research purposes. For intrinsic image decomposition, the \emph{clean pass} images and the corresponding albedo and shading layers have been published as the MPI Sintel dataset, containing 18 sequences with a total of 890 frames. As discussed in \ref{sec:data_refine}, there is severe illumination inconsistency between the input frames and the shading layers in this dataset. Therefore, we provide the refined MPI Sintel dataset as a more suitable dataset for intrinsic image decomposition.

Figure \ref{fig:4}(bottom) compares our refined MPI dataset (MPI\_RD) to the original MPI dataset (MPI).
In the shading layer of the first column, we can see that in our refined shading image, the specularity on the shoulder of the girl is removed, making the shading illumination consistent with the original input image.
In the second and third columns, the shading layers from the MPI\_RD dataset contain more geometric details than those from the MPI dataset. For instance, the wooden cart's coarse surface is depicted in the refined shading in the third column, while the original shading from the MPI dataset  has a smooth surface. These examples demonstrate that our refined MPI\_RD improves consistency between the intrinsic decomposition and the input image.
In Figure \ref{fig:4}(bottom, right), the mean squared error (MSE) between the input image $I$ and the resynthesized image $A\times S$ is computed. The MSE for the MPI\_RD dataset is significantly smaller than for the MPI dataset, showing that the intrinsic decomposition model Eq. (\ref{intrinsic}) is well respected in the refined dataset.

\begin{table*}[!t]  % table* for inserting 2-collum table
    %	\scriptsize
    
%    \small
    \renewcommand{\arraystretch}{1.2}
    \caption{Comparison of methods  using the MPI Sintel dataset.}
    \label{tab:1}
%    \label{table_MPI}
    
    \centering
    %	\setlength\tabcolsep{0.5pt}
%    \resizebox{0.5\textwidth}{!}{
        \begin{tabular}{L{2mm} l rrr rrr rrr}
            \toprule
            &\ &\multicolumn{3}{c}{MSE} &\multicolumn{3}{c}{LMSE} &\multicolumn{3}{c}{DSSIM} \\
            \cmidrule(lr){3-5} \cmidrule(lr){6-8} \cmidrule(lr){9-11}
            &Methods  & albedo  & shading  & avg  & albedo  & shading  & avg  & albedo  & shading  & avg \\
            \midrule
            \multirow{6}{*}{\rotatebox{90}{\textit{image split}}}
            &retinex \cite{grosse2009-dataset} & 0.0606 & 0.0727 & 0.0667 & 0.0366  & 0.0419  & 0.0393 & 0.2270 & 0.2400 & 0.2335 \\
            &Barron et al. \cite{barron2014-shape} & 0.0420 & 0.0436 & 0.0428  & 0.0298  & 0.0264  & 0.0281 & 0.2100 & 0.2060 & 0.2080 \\
            &Chen et al. \cite{chen2013-depth}  & 0.0307 & 0.0277 & 0.0292  & 0.0185  & 0.0190  & 0.0188 & 0.1960 & 0.1650 & 0.1805 \\
            &MSCR \cite{NMY2015-DI}   & 0.0100 & 0.0092 & 0.0096  & 0.0083  & 0.0085  & 0.0084 & 0.2014 & 0.1505 & 0.1760 \\
            &Revisiting \cite{fan2018-revisiting}  & 0.0069 & 0.0059 & 0.0064  & 0.0044  & 0.0042  & 0.0043 & 0.1194 & 0.0822 & 0.1008 \\
            &Ours    & \textbf{0.0047} & \textbf{0.0046} & \textbf{0.0047}  & \textbf{0.0037}  & \textbf{0.0038}  & \textbf{0.0038} & \textbf{0.0950} & \textbf{0.0774} & \textbf{0.0862} \\
            \midrule
            \multirow{3}{*}{\rotatebox{90}{\textit{scene split}}}
            &MSCR \cite{NMY2015-DI}   & 0.0190 & 0.0213 & 0.0201  & 0.0129  & 0.0141  & 0.0135 & 0.2056 & 0.1596 & 0.1826 \\
            &Revisiting \cite{fan2018-revisiting}  & 0.0189 & \textbf{0.0171} & \textbf{0.0180}  & 0.0122  & \textbf{0.0117}  & \textbf{0.0119} & 0.1645 & 0.1450 & 0.1547 \\
            &Ours & \textbf{0.0173} & 0.0195 & 0.0184  & \textbf{0.0118}  & 0.0147  & 0.0133 & \textbf{0.1587} & \textbf{0.1405} & \textbf{0.1496} \\
            \bottomrule
        \end{tabular}
%    }
\end{table*}

For data augmentation, we randomly resize the input image by a scale factor in $[0.8,~1.3]$, and randomly crop a $288\times 288$ patch from the resized image per iteration. We also use horizontal flipping in the training phase. When comparing methods, following \cite{fan2018-revisiting}, we evaluate our results on both a scene split and an image split. For a scene split, half of the scenes are used for training and the other half for testing. For an image split, all 890 images are randomly separated into two sets. Evaluation on a scene split is considered more challenging as it requires more generalization capacity.
    
    For video data, the refined version MPI\_VRD \emph{(video refined data)} alleviate image illumination jitter and  local region inconsistencies. Quantitatively, taking the shading images of video sleeping\_2 for example,  temporal consistency refinement increases TCM  from 0.74 to 0.78.

\subsubsection{IIW dataset}
Intrinsic Images in the Wild (IIW) \cite{bell2014-dataset} is a large scale, public dataset of real-world scenes intended for intrinsic image decomposition. It contains 5,230 real images of mostly indoor scenes, combined with a total of 872,161 crowd-sourced annotations of reflectance comparisons between pairs of points sparsely selected throughout the images (on average 100 judgements per image). Following many prior works \cite{narihira2015-relative, nestmeyer2017-filtering, zhou2015-datadriven, fan2018-revisiting}, we split the IIW dataset by placing the first of every five consecutive images sorted by image ID into the test set, and the others into the training set.  WHDR from \cite{bell2014-dataset} is employed to measure the quality of the reconstructed albedo images.

For the IIW dataset, our proposed network structure cannot be directly used due to the lack of dense labelling of albedo and shading layers. Actually, only sparse and relative reflectance annotations are provided. In order to take advantage of the proposed feature distribution divergence and feature distribution consistency, we  modify the network. In detail, the predicted dense albedo is collected into an image pool to describe the distribution of albedo. The reconstructed shading using the original image and predicted albedo is used as  dense supervision for  shading prediction, and is also collected in an image pool to describe the shading distribution. We set the weights in Eq.~(\ref{feature_distribution_constraint}) to  $[\gamma_1,\dots,\gamma_5] = [0,~0,~0,~1.0,~1.0]$.

%%% Qualitative Evaluation
\subsection{Comparison to state-of-the-art methods}

\subsubsection{Using MPI Sintel and the refined dataset}

\begin{figure*}[!tp]
    \centering
    \includegraphics[width=\linewidth]{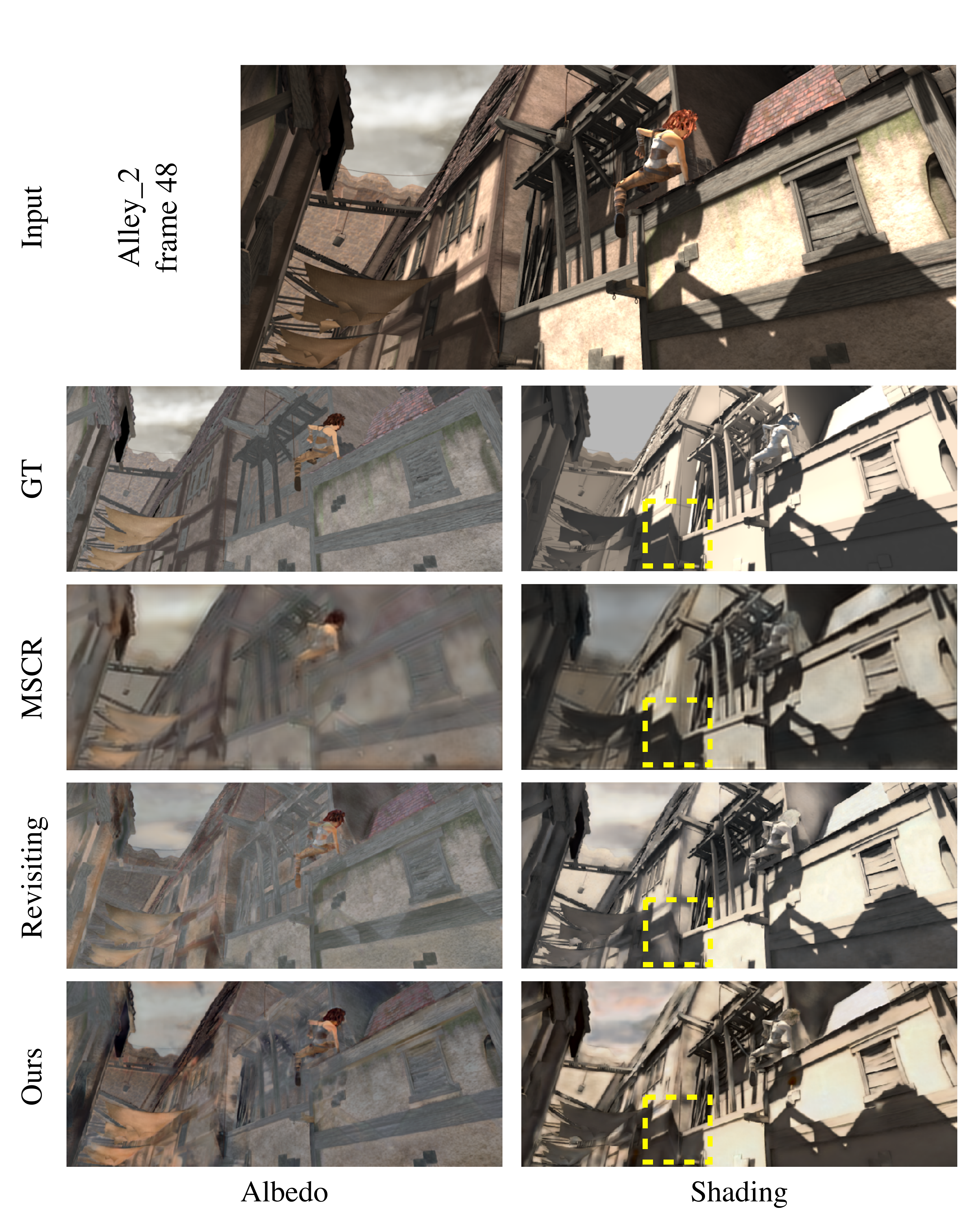}
    \caption{Qualitative comparison on the MPI Sintel dataset. The visual results are evaluated on the more challenging scene split. The regions with obvious differences are highlighted in dotted boxes.}
    \label{fig:5}
%    \label{MPI_results}
\end{figure*}

As Table \ref{tab:1} shows, our method achieves the best results on the MPI Sintel dataset using the image split. On the more challenging scene split, our method is competitive with the state of the art, and achieves the best results in 5 out of the 9 cases in the table. We show a group of qualitative results evaluated on the scene split in Figure \ref{fig:5}. While the MSCR \cite{NMY2015-DI} results are relatively blurred due to the large kernel convolutions and down-sampling, our method provides sharper results comparable to Revisiting \cite{fan2018-revisiting}. Moreover, our shading layer depicts better shadow area than \cite{fan2018-revisiting}.

As explained in \ref{sec:data_refine}, the MPI Sintel dataset has issues of data consistency between the original input images and the corresponding shading images. Because of the proposed feature distribution divergence, feature distribution consistency and the use of the cycle loss, our method is sensitive to such data inconsistency.
Therefore, we compare our method to the state-of-the-art methods on the more challenging scene split of the refined MPI Sintel dataset. As Table \ref{tab:2} shows, our method achieves the best result,  demonstrating the effectiveness of our method and data refinement process.

\begin{table*}[!t]  % table* for inserting 2-collum table
    %	\scriptsize
   \renewcommand{\arraystretch}{1.2}
    \caption{Comparison of methods  using the MPI\_RD dataset.}
    \label{tab:2}
%    \label{table_RD_MPI}
    
    \centering
    %	\setlength\tabcolsep{0.5pt}
    %	\begin{threeparttable}
            \begin{tabular}{L{2mm} l rrr rrr rrr}
                \toprule
                &\ &\multicolumn{3}{c}{MSE} &\multicolumn{3}{c}{LMSE} &\multicolumn{3}{c}{DSSIM} \\
                \cmidrule(lr){3-5} \cmidrule(lr){6-8} \cmidrule(lr){9-11}
                &Methods  & albedo  & shading  & avg  & albedo  & shading  & avg  & albedo  & shading  & avg \\
                \midrule
%                \multirow{2}{*}%{\textit{scene split}}
                &MSCR \cite{NMY2015-DI}        & 0.0222 & 0.0175 & 0.0199  & 0.0151  & 0.0122  & 0.0136 & 0.1803 & 0.1619 & 0.1711 \\
                &Revisiting \cite{fan2018-revisiting}  & 0.0196 & 0.0137 & 0.0167  & 0.0146  & 0.0094  & 0.0120 & 0.1651 & 0.1082 & 0.1366 \\
                
                \midrule
                \multirow{7}{*}{\rotatebox{90}{\textit{ablation studies}}}
                &Plain w/o ssim or grad
                &0.0189 &0.0147 &0.0168
                &0.0137 &0.0104 &0.0120
                &0.1560 &0.1140 &0.1350 \\
                &Plain w/o ssim
                &0.0169 &0.0149 &0.0159
                &0.0117 &0.0103 &0.0110
                &0.1530 &0.1100 &0.1320 \\
                &Plain w/o grad
                &0.0188 &0.0142 &0.0165
                &0.0123 &0.0095 &0.0109
                &0.1520 &0.1090 &0.1300 \\
                
                &Plain    & 0.0172 & 0.0147 & 0.0159  & 0.0116  & 0.0097  & 0.0106 & 0.1528 & 0.1085 & 0.1307 \\
                &Ours w/o FDD            & 0.0166 & 0.0134 & 0.0150  & 0.0112  & 0.0090  & 0.0101 & 0.1474 & 0.1048 & 0.1261 \\
                &Ours w/o FDC          & 0.0170 & 0.0130 & 0.0150  & 0.0113  & 0.0089  & 0.0101 & 0.1530 & 0.1070 & 0.1300 \\
                &Ours                   & \textbf{0.0157} & \textbf{0.0126} & \textbf{0.0142}  & \textbf{0.0105}  & \textbf{0.0087}  & \textbf{0.0096} & \textbf{0.1419} & \textbf{0.1015} & \textbf{0.1217} \\
                \bottomrule
            \end{tabular}
        % 		\begin{tablenotes}
            % 			\begin{small}
                % 				\footnotesize
                % \item[] (`Ours plain' is the basic two-stream network without FDD or FDC.)
                % 			\end{small}
            % 		\end{tablenotes}
        %	\end{threeparttable}
    
\end{table*}

\begin{figure*}[!tp]
    \centering
    \includegraphics[width=\linewidth]{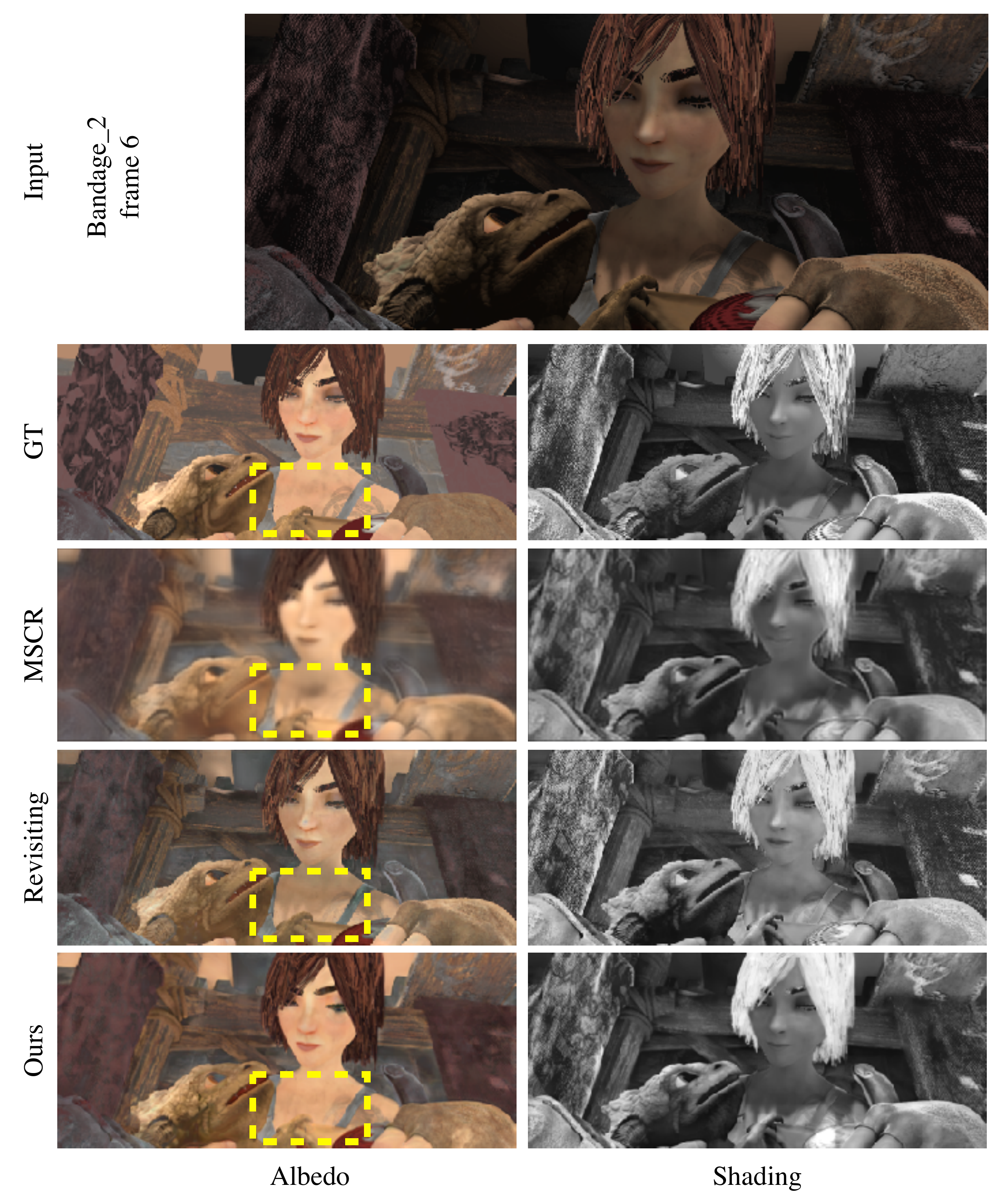}\vspace {-1mm}
    \caption{Qualitative comparison on our refined MPI Sintel dataset MPI\_RD. Visual results are evaluated on the scene split. Our method is better at separating albedo and shading components. The regions with obvious difference are highlighted in dotted boxes.}
    \label{fig:6}
    %    \label{MPI_results_2}
\end{figure*}

\begin{table*}[!h]  % table* for inserting 2-collum table
    %	\scriptsize
    \renewcommand{\arraystretch}{1.2}
    \caption{Method comparison and alternative training approach comparison using the MIT intrinsic dataset. The top 3 methods are classical methods, while the others are learning based ones. Note that Barron et al.'s method relies on specialized priors and masked objects particular to this dataset. Alternative training strategies are given in the bottom 3 rows.  Ours (scratch) means training from scratch. Ours (MPI) means pre-trained on the original MPI Sintel dataset. Ours (RD) means pre-trained on the refined MPI Sintel dataset.  }
    \label{tab:3}
%    \label{table_MIT}
    
    \centering
            \begin{tabular}{ l ccc c}
                \toprule
                \ &\multicolumn{3}{c}{MSE} &\multicolumn{1}{c}{LMSE} \\
                \cmidrule(lr){2-5}
                Method  & albedo  & shading  & avg  & total \\
                \midrule
%                \multirow{3}{*}%{\textit{scene split}}
                Barron et al. \cite{barron2014-shape}        &0.0064 &0.0098 &0.0081  &0.0125   \\
                SRIE \cite{fu2016srie}                                 &0.0136 &0.0128 &0.0132  &0.0192 \\
                Relative Smoothness \cite{li2014-smooth}   &0.0492 &0.0342 &0.0417  &0.0216 \\
                \midrule
%                \multirow{4}{*}{\rotatebox{90}{\textit{learning based}}}
                Zhou et al. \cite{zhou2015-datadriven}          &0.0252 &0.0229 &0.0240  &0.0319   \\
                Shi et al. \cite{shi2017-nonlambertian}           &0.0216 &0.0135 &0.0175  &0.0271   \\
                MSCR \cite{NMY2015-DI}                 &0.0207 &0.0124 &0.0165  &0.0239   \\
                Revisiting \cite{fan2018-revisiting}      &0.0134 &\textbf{0.0089} &0.0111  &0.0203   \\
                \midrule
                Ours (scratch)             & 0.0134 &0.0099 &0.0117 &0.0186\\
                Ours (MPI)                 & 0.0126 &0.0106 &0.0116 &0.0175\\
                Ours (RD)                  & \textbf{0.0120} & 0.0095 & \textbf{0.0108}  & \textbf{0.0170} \\
                \bottomrule
            \end{tabular}
\end{table*}

    To further validate the effectiveness of the proposed method, we also conducted an ablation study on the training loses as well as the network architecture. Results  are given in the bottom part of Table \ref{tab:2}. `Plain' represents the baseline two-stream network structure shown in
    Figure \ref{fig:2}(b). In the ablation study, the gradient loss and SSIM loss are progressively added to train the plain network. The experimental results show that using the gradient loss and SSIM loss simultaneously achieves better intrinsic image decomposition results.
    The proposed network architecture is denoted by `Ours', as defined in Figure \ref{fig:2}(c).
    It can be observed that using only the feature distribution divergence (w/o FDC) or the feature distribution consistency (w/o FDD) does not improve results much, while using both of them results in considerable improvement.

Figure \ref{fig:6} displays a side-by-side comparison with two other methods using the refined dataset MPI\_RD. As can be seen, our method  better  separates shading from albedo information. For example, our method outputs consistent shadow  around the girl's neck.

\begin{figure*}[!h]
    \centering
    \includegraphics[width=\linewidth]{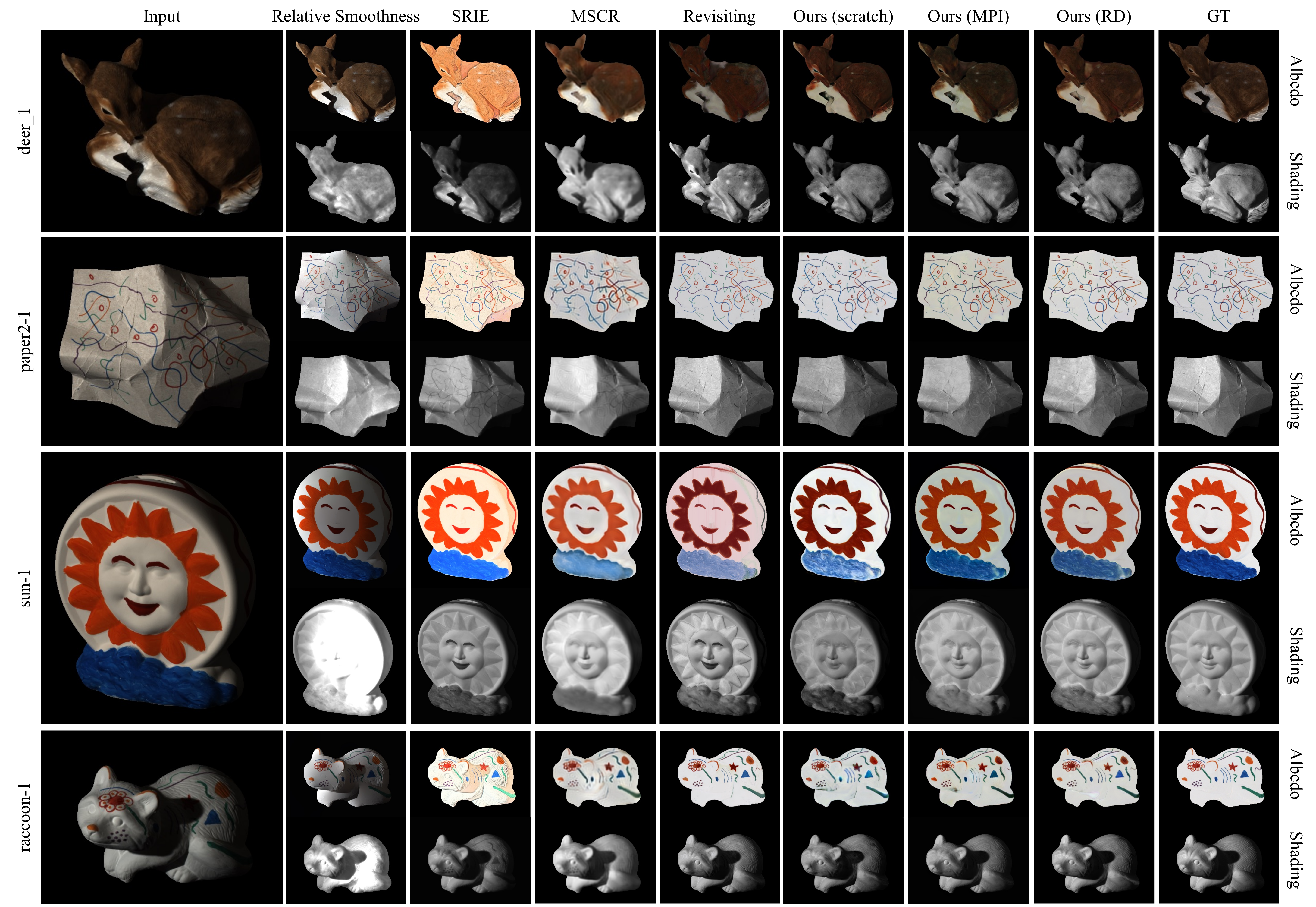}
    \caption{Qualitative comparison on the MIT intrinsic dataset. The intrinsic image decomposition results from classical methods (columns 2 and 3) and learning based methods (columns 4 and 5) are compared with our result (column 8). Results of different training strategies for our network (columns 6--8) are also provided.}
    \label{fig:7}
%    \label{MIT_results}
\end{figure*}

\subsubsection{Using the MIT intrinsic dataset}

We further experimented on the MIT intrinsic dataset \cite{grosse2009-dataset}, which consists of object-level real images. 
In this experiment, classical methods \cite{barron2014-shape,fu2016srie,li2014-smooth} as well as learning based methods \cite{fan2018-revisiting, shi2017-nonlambertian, NMY2015-DI, zhou2015-datadriven} were compared to our proposed method.
We also conduct an ablation study on the training strategies of our proposed network, including training from scratch (Ours scratch), pre-training on the original MPI Sintel dataset (Ours MPI), and pre-training on the refined MPI Sintel dataset (Ours RD). The experimental results are reported in Table \ref{tab:3}, and representative instances are selected for visual comparison in Figure \ref{fig:7}.

As in \cite{fan2018-revisiting}, we used the 220 images in the dataset. In comparisons to previous methods, the split from \cite{barron2014-shape} was used.
Our refined MPI Sintel dataset has grayscale shading images, so we first pre-train the model on  MPI\_RD and then fine-tuned it on the MIT training set.

Numerical results are shown in Table \ref{tab:3}. Our method (Ours RD) achieves the best results in most cases in the table.
Moreover,  Ours RD performs better than Ours MPI in terms of LMSE, and both pre-training methods perform better than training from scratch, demonstrating that pre-training on the refined MPI Sintel dataset helps intrinsic image decomposition on the MIT dataset.

Qualitative results are illustrated in Figure \ref{fig:7}. We can observe that our method (Ours RD) predicts sharp and accurate intrinsic layers.
Classical methods may produce meaningful layer separation results, but good results depend on parameter tuning, which requires expert knowledge. Compared to (Ours scratch) and (Ours MPI), (Ours RD) achieves better region consistency with the ground truth.

\begin{figure*}[!th]
    \centering
    \includegraphics[width=1\linewidth]{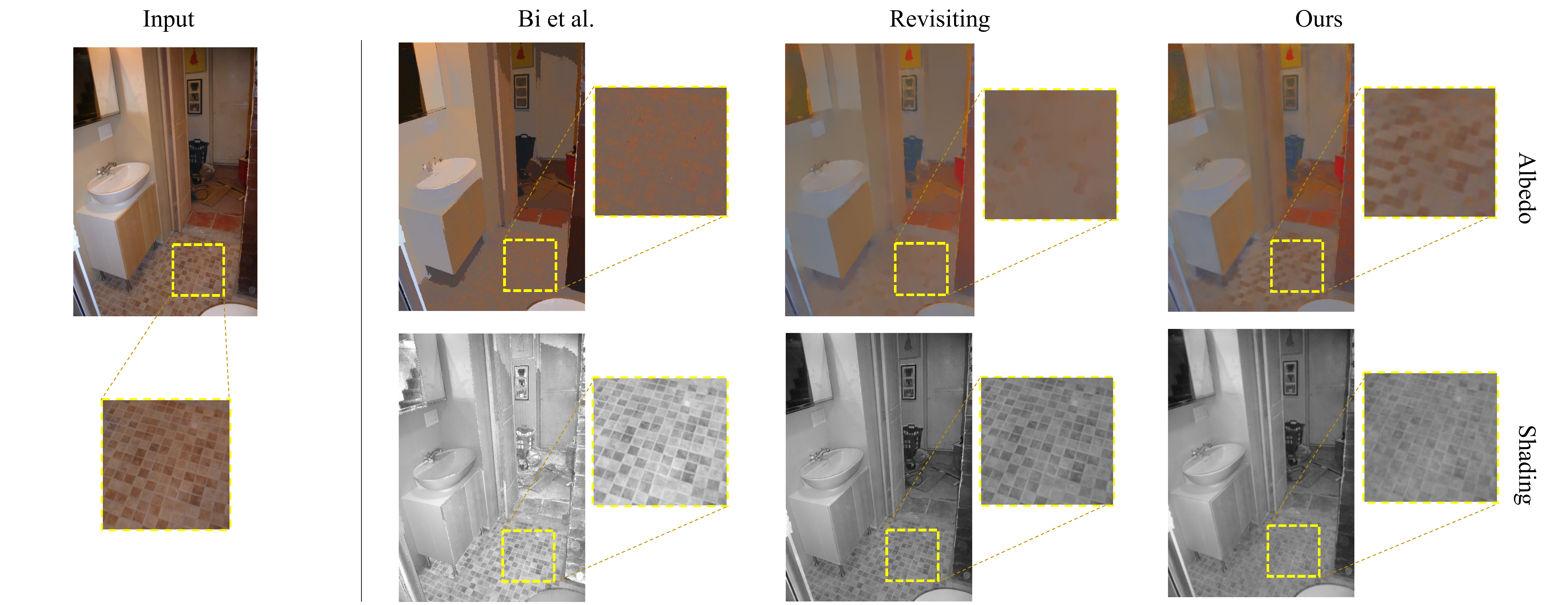}
    \caption{Qualitative method comparison on the IIW dataset. Regions with obvious difference are highlighted in dotted boxes,  also shown enlarged.}
    \label{fig:8}
    %    \label{IIW_results}
\end{figure*}

\subsubsection{Using the IIW dataset} \label{sec:iiw_experiments}

\begin{table}[!t]  % table* for inserting 2-collum table
    %	\scriptsize
%    \small
%    \renewcommand{\arraystretch}{1.2}
%    \vspace {-2.5mm}
    \caption{Method comparison on the IIW test set.}
    \label{tab:4}
%    \label{table_IIW}
    
    \centering
    %	\setlength\tabcolsep{0.5pt}
%    \resizebox{0.42\textwidth}{!}{
        \begin{tabular}{ l r}
            \toprule
            Method  & WHDR (mean) \\
            \midrule
            Baseline (const shading) &51.37 \\
            Baseline (const reflectance) &36.54 \\
            Shen et al. 2011 \cite{shen2011-sparsity}  &36.90 \\
            Retinex (color) \cite{grosse2009-dataset}  &26.89 \\
            Retinex (gray)  \cite{grosse2009-dataset}  &26.84 \\
            Garces et al. 2012 \cite{garces2012-clustering}  &25.46 \\
            Zhao et al. 2012 \cite{zhao2012-nonlocal_texture} &23.20 \\
            $L_1$ flattening  \cite{sai2015-L1Intrinsic}  &20.94 \\
            Bell et al. 2014 \cite{bell2014-dataset}  &20.64 \\
            Zhou et al. 2015 \cite{zhou2015-datadriven} &19.95 \\
            Nestmeyer et al. 2017 (CNN) \cite{nestmeyer2017-filtering}  &19.49 \\
            Zoran et al. 2015* \cite{zoran2015-ordinal}  &17.85 \\
            Nestmeyer et al. 2017 \cite{nestmeyer2017-filtering}  &17.69 \\
            Bi et al. 2015 \cite{sai2015-L1Intrinsic}  &17.67 \\
            CGIntrinsic \cite{li2018-cgintrinsics} &14.80 \\
            Revisiting \cite{fan2018-revisiting}  &14.45 \\
            \midrule
            Ours  & \textbf{13.60} \\
            \bottomrule
        \end{tabular}
%    }
\end{table}

In Table \ref{tab:4}, we report  results obtained using the test set of the IIW dataset. Our proposed method achieves the best performance with a mean WHDR value of $13.60\%$,  a considerable improvement over the second best \cite{fan2018-revisiting} with a mean WHDR value of $14.45\%$. To better show the quality of our results, we give a qualitative comparison to  state-of-the-art methods in Figure \ref{fig:8}.
Detailed intrinsic decomposition results are shown in the close-up windows. It can be observed that our method successfully preserves the texture of the floor tiles in the albedo layer, while the other approaches treat such texture as shading.

\begin{figure*}[h]
    \centering
    \includegraphics[width=1.0\linewidth]{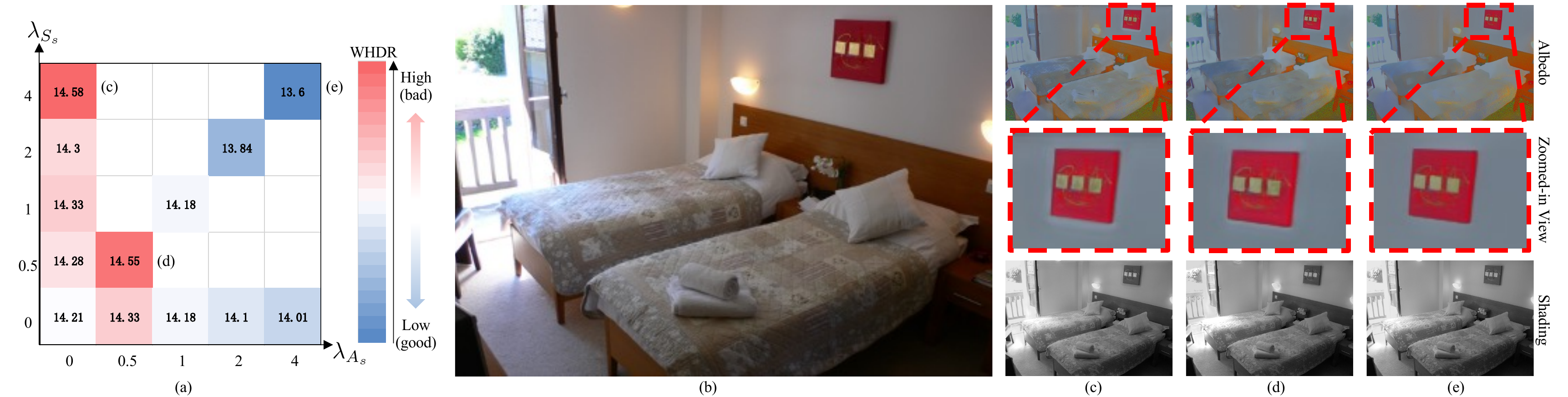}
    \caption{Quantitative (a) and qualitative  (b--e) results of varying smoothness priors. (a): WHDR scores with varying albedo smoothness weight $\lambda_{A_s}$ (horizontal axis) and  shading smoothness weight $\lambda_{S_s}$ (vertical axis) from 0 to 4. Numerical values and color coding are provided, with red being worse. (b): input image. (c---e): intrinsic decomposition images for different weight settings: $(\lambda_{A_s}, \lambda_{S_s})=(0, 4), (0.5, 0.5), (4, 4)$, respectively. Regions with obvious differences are highlighted in red boxes.
    }
    \label{fig:9}
    %    \label{fig:iiwAblation}
\end{figure*}

    To further investigate the influence of the smoothness priors applied in the sparsely-labelled case, a comparison was conducted.  Figure \ref{fig:9} shows the effects of changing the weights of the smoothness terms in Eq. (\ref{total_loss_sparse}): the albedo smoothness weight $\lambda_\mathrm{As}$ and shading smoothness weight $\lambda_\mathrm{Ss}$ were varied from 0 to 4 respectively. Results show that using the shading smoothness term increases the reconstruction loss, while using the albedo smoothness term decreases the reconstruction loss. Using both simultaneously  provides the best result. In Figure \ref{fig:9}(c, d, e), three representative parameter settings  are used to show the resulting intrinsic decomposition. The albedo image in Figure \ref{fig:9}(e) contains more precise contours and consistent region colors.

    \subsubsection{Intrinsic decomposition of video data}
    In this experiment, we evaluated the proposed intrinsic decomposition method on video data with respect to reconstruction quality of the intrinsic images, and to temporal consistency, through comparisons to alternative methods.
    
    Framewise methods used for comparison include MSCR \cite{NMY2015-DI} and Revisiting \cite{fan2018-revisiting}. 
    The blind video temporal consistency method \cite{lei2020blind} is an unsupervised video smoothing method. In the experiment (Ours+DVP), it is applied as post-processing to increase the temporal consistency of the results produced by our framewise method (Ours). (Ours+Flow) is the proposed intrinsic decomposition method for video data. The temporal consistency constraint is applied by taking optical flow as input to construct correspondences between adjacent frames. We also extend the temporal consistency constraint to MSCR in the same way as for (Ours+Flow) in (MSCR+Flow).
    All methods were trained and tested on the MPI\_VRD dataset.
    
    The intrinsic decomposition accuracy scores of framewise methods and flow methods are shown in Table \ref{tab:6}, while the temporal consistency  scores for 8 test short videos are shown in Table \ref{tab:7}.
    Using adjacent frame temporal consistency as a constraint, (Ours+Flow) and (MSCR+Flow) achieve much better temporal consistency metric scores than their framewise counterparts, while the intrinsic decomposition accuracy scores remain almost unchanged. This demonstrates the effectiveness of our proposed extension method for video data. As a result, (Ours+Flow) achieves the best average accuracy and temporal consistency for intrinsic decomposition of video data.
    
\begin{figure*}[!ht]
        \centering
        \includegraphics[width=1.0\linewidth]{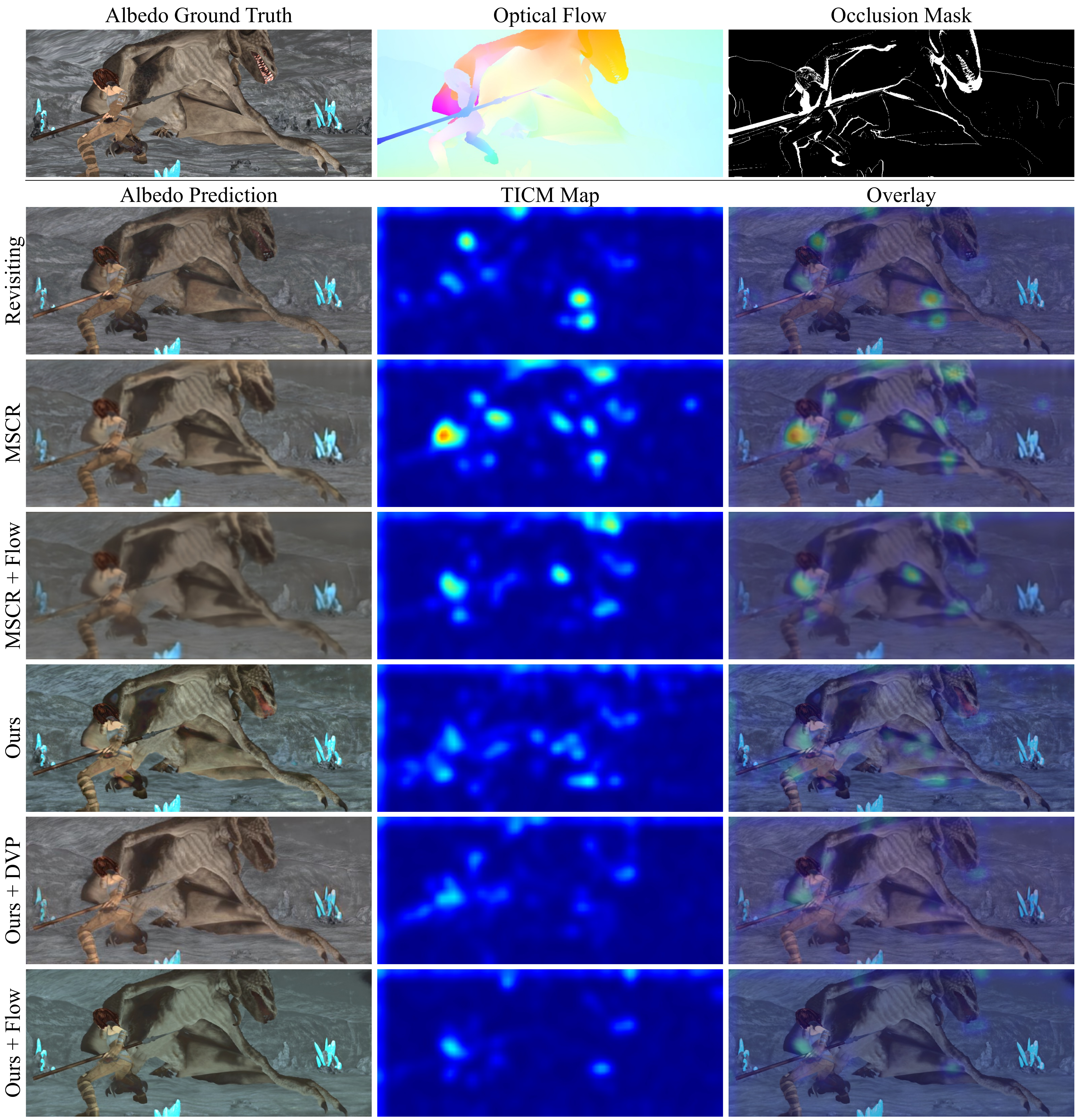}
        \caption{Qualitative comparison of different intrinsic decomposition methods on video data. Left to right, top row: albedo ground truth, optical flow map, and  occlusion mask (pixels in white have invalid optical flow). Rows 2--6:  intrinsic decomposition results for various methods, showing albedo prediction, TICM map, and both overlayed. Colder  colors indicate better  temporal consistency. By using temporal consistency loss, both MSCR+Flow and Ours+Flow achieve better temporal consistency, compared to their framewise counterparts MSCR and Ours. Results show frame 32 in cave\_4.
        }
        \label{fig:10}
        %    \label{fig:tcm_show}
    \end{figure*}

    We also visualize temporal consistency of a specific frame clipped from a video, using a \emph{temporal inconsistency metric} (TICM) map to highlight temporally inconsistent areas. 
    First, the TCM map is computed using Eq. (\ref{Eq_tcmmap}). Then, the TCM map is smoothed by a Gaussian kernel of size 65. Finally, the Jet colormap is inverted to highlight inconsistent areas with warm colors (the colder the color, the better the consistency).
A qualitative comparison of different intrinsic decomposition methods on video data is shown in Figure \ref{fig:10}.  It can be observed that, using the temporal consistency constraint, both (MSCR+Flow) and (Ours+Flow) achieve better temporal consistency compared to their framewise counterparts (MSCR and Ours).
    
\begin{figure*}[!ht]
    \centering
    \includegraphics[width=1.0\linewidth]{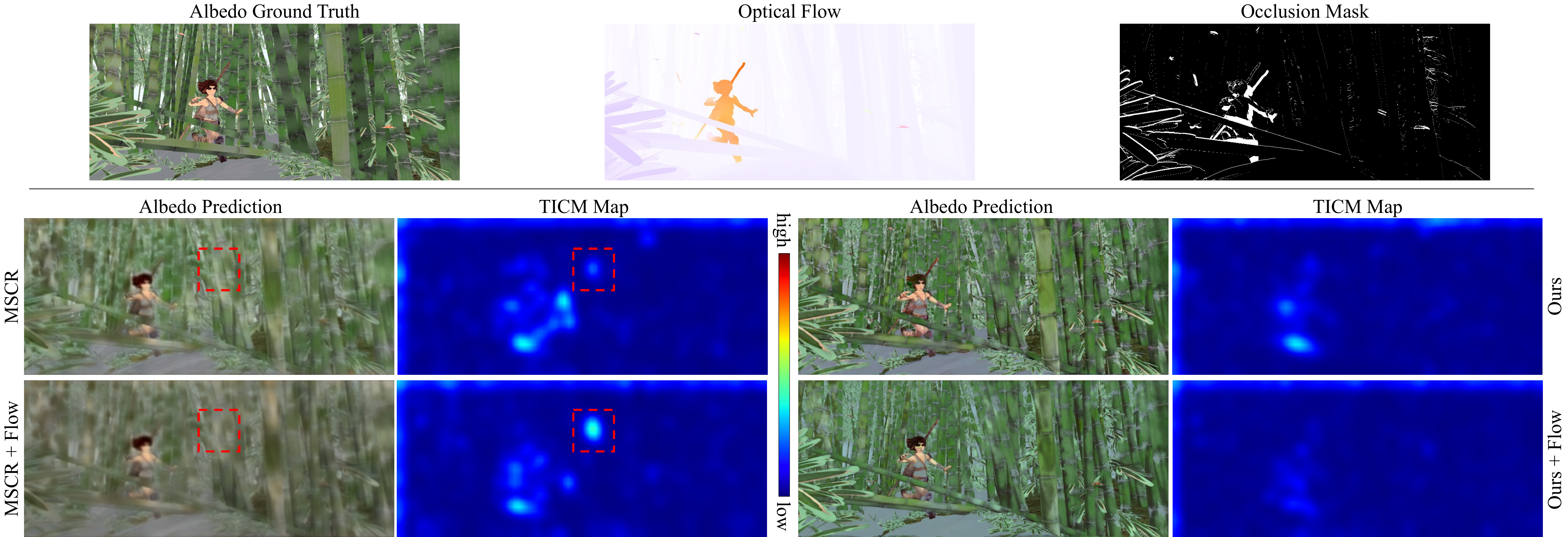}
    \caption{Effects of using temporal consistency loss, showing frame 23 in bamboo\_2. Temporal consistency loss is added to the MSCR method (bottom left) and our method (bottom right). In the Ours+Flow result, the temporally inconsistent areas are suppressed compared to the Ours result. However, in the MSCR+Flow result, temporal inconsistency is unexpectedly amplified in the highlighted area in the red box. In addition, the MSCR+Flow result is more blurred.
    }
    \label{fig:11}
    %    \label{fig:tcm_discuss}
\end{figure*}

In Figure \ref{fig:11}, temporal consistency loss is added to framewise methods (MSCR) and (Ours). In the (Ours+Flow) result, temporally inconsistent areas are suppressed compared to (Ours). However, in the (MSCR+Flow) result, temporal inconsistency is unexpectedly amplified in the highlighted area in the red box. In addition, the (MSCR+Flow) result is more blurred than the one without  temporal consistency loss. In MSCR, multiple scales of feature maps are merged in the encoding phase. While  features from deeper layers contain high-level knowledge, they may lack texture details. The temporal consistency constraint may encourage MSCR to pay more attention to high-level features,  resulting in more blurred outputs.
    The case of (MSCR+Flow) indicates that while temporal consistency loss could be easily extended to other deep neural networks, the outcome will largely depend on the characteristics of specific methods.

    \begin{table*}[!ht]  % table* for inserting 2-collum table
        %	\scriptsize
%        \small
        \renewcommand{\arraystretch}{1.2}
        \caption{Method comparison for intrinsic decomposition of video data.}
        \label{tab:6}
%        \label{table_MPI_VRD}
        
        \centering
        %	\setlength\tabcolsep{0.5pt}
%        \resizebox{0.5\textwidth}{!}{
            \begin{tabular}{L{0mm} l rrr rrr rrr}
                \toprule
                &\ &\multicolumn{3}{c}{MSE} &\multicolumn{3}{c}{LMSE} &\multicolumn{3}{c}{DSSIM} \\
                \cmidrule(lr){3-5} \cmidrule(lr){6-8} \cmidrule(lr){9-11}
                &Methods  & albedo  & shading  & avg  & albedo  & shading  & avg  & albedo  & shading  & avg \\
                \midrule
                \multirow{3}{*}{\rotatebox{90}{\textit{framewise}}}
                &MSCR \cite{NMY2015-DI}
                    & 0.0223 &0.0181 &0.0202
                    & 0.0154 &0.0127 &0.0140
                    & 0.1880 &0.1609 &0.1745 \\
                &Revisiting \cite{fan2018-revisiting}
                    & 0.0199 & 0.0137 & 0.0168
                    & 0.0148  & 0.0094  & 0.0121
                    & 0.1655 & 0.1095 & 0.1375 \\
                &Ours
                    &0.0173  &0.0163  &0.0168
                    &0.0120  &0.0115  &0.0117
                    &0.1472  &0.1123  &0.1298 \\
                \midrule
                \multirow{3}{*}{\rotatebox{90}{\textit{flow}}}
                &MSCR+Flow
                    &0.0227	&0.0184	&0.0206	
                    &0.0150	&0.0123	&0.0136	
                    &0.1950	&0.1727	&0.1839 \\
                &Ours+DVP \cite{lei2020blind}
                    & 0.0169 & 0.0165 & 0.0167
                    & 0.0117  & 0.0114  & 0.0116
                    & 0.1549 & 0.1204 & 0.1377 \\
                
                &Ours+Flow
                    & 0.0173 & 0.0152 & \textbf{0.0163}
                    & 0.0117  & 0.0102  & \textbf{0.0109}
                    & 0.1423 & 0.1054 & \textbf{0.1238} \\
                
                \bottomrule
            \end{tabular}
     \end{table*}

    \begin{table*}[!ht]  % table* for inserting 2-collum table
        %	\scriptsize
%        \small
        \renewcommand{\arraystretch}{1.2}
        \caption{Temporal consistency values for various methods for intrinsic decomposition of video ($\times 10^{-2}$).
            A to H represent the videos:
            alley\_2, bamboo\_2, bandage\_2, cave\_4,
            market\_5, mountain\_1, sleeping\_2, temple\_3.}
        \label{tab:7}
%        \label{table_VRD_TCM}
        
        \centering
        %	\setlength\tabcolsep{0.5pt}
%        \begin{threeparttable}
                \begin{tabular}{L{0mm} l rrr rrr rrr}
                    \toprule
                    &Methods  & A  & B  & C  & D  & E  & F  & G  & H  & \textbf{avg} \\
                    \midrule
                    \multirow{3}{*}{\rotatebox{90}{\textit{framewise}}}
                    &MSCR \cite{NMY2015-DI}
                    &48.18	&62.30	&38.70	&31.80	&5.49	&52.36	&86.50	&31.47	&44.60
                    \\
                    &Revisiting \cite{fan2018-revisiting}
                    &8.32	&35.48	&18.06	&25.44	&5.03	&68.92	&19.37	&39.41	&27.50
                    \\
                    &Ours
                    &0.12	&27.61	&6.55	&4.57	&2.04	&88.99	&12.80	&40.05	&22.84
                    \\
                    \midrule
                    \multirow{3}{*}{\rotatebox{90}{\textit{flow}}}
                    &MSCR+Flow
                    &68.69	&56.42	&50.50	&56.76	&7.87	&48.27	&60.34	&31.78	&47.58
                    \\
                    
                    &Ours+DVP \cite{lei2020blind}
                    &78.93	&47.52	&28.72	&10.19	&22.71	&51.89	&72.33	&54.06	&45.79
                    \\
                    
                    &Ours+Flow
                    &84.22	&70.91	&78.33	&60.32	&6.42	&51.64	&54.05	&45.12	&\textbf{56.37}
                    \\
                    
                    \bottomrule
                \end{tabular}
%            \begin{tablenotes}
%                \begin{small}
%                    \footnotesize
%                    \centering
%                    \item[] (A to H represents 8 videos:
%                    % 		\item[]
%                    alley\_2, bamboo\_2, bandage\_2, cave\_4,
%%                    \item[]
%                    market\_5, mountain\_1, sleeping\_2, temple\_3.)
%                \end{small}
%            \end{tablenotes}
%        \end{threeparttable}
    \end{table*}

\section{Conclusions}\label{sec:conclusions}

    In this paper, we have presented a novel two-stream encoder-decoder network for intrinsic image decomposition. Our method is able to exploit  discriminative properties of the features for different intrinsic images. Specifically, our feature distribution divergence is designed to increase the distance between features corresponding to different intrinsic images, and our feature perceptual loss is applied to constrain the feature distribution. These two modules work together to encode discriminative features for intrinsic image decomposition. We have also provided an algorithm to refine the MPI Sintel dataset to make it more suitable for intrinsic image decomposition.
    Visual results for  MPI\_RD and the more challenging IIW dataset demonstrate that our proposed method can achieve results with better albedo and shading separation than existing methods.
    Its extension to video data is able to decompose video into intrinsic image sequences with temporal consistency.
    
    The limitations of our method are of two kinds. Firstly, the mathematical model of intrinsic decomposition applied in this work is relatively preliminary. We cannot recover scene properties such as 3D geometry, light source positions, global illumination, etc. It is worthwhile challenge to exploit  feature discrimination properties in more complex and powerful mathematical models.
    Secondly, the temporal consistency constraint used to extend our method to  intrinsic video decomposition does not generalize well. As Figure \ref{fig:11} shows, directly using the temporal consistency loss in MSCR can result in unwanted blurring. In future, it is of interest to derive more general temporal consistency constraints for intrinsic video decomposition.

%============================================%

\noindent\textbf{Acknowledgements.}
Portions of this work were presented at the International Conference on Computer Vision Workshops in 2019 \cite{wang2019single}.
This work was supported by the National Natural Science Foundation of China (NSFC)  (Grants 61972012, 61732016).

\noindent\textbf{Declarations}

\textbf{Conflict of interest} The authors declare that they have no conflict of interest.

\appendices

\ifCLASSOPTIONcaptionsoff
  \newpage
\fi

\section{Constraints for sparsely-labelled data}
\subsection{Ordinal loss}

For each pair of annotated pixels $(i,j)$ in the predicted albedo image $A$, we have the error function:
    \begin{equation}
        e_{i,j}(A) =\left\{
   {\small
        \begin{aligned}
            &\omega_{i,j}(\log~A_i - \log~A_j)^2,         & r_{i,j}&=0  \\
            &\omega_{i,j}(\max(0, m-\log~A_i+\log~A_j))^2, & r_{i,j}&=+1 \\
            &\omega_{i,j}(\max(0, m-\log~A_j+\log~A_i))^2, & r_{i,j}&=-1
        \end{aligned}
    }
        \right.
        \label{ordinal_error}
    \end{equation}
where $r_{i,j}$ denotes the relative reflectance (albedo) judgement from  IIW. Values of $r_{i,j}$ are set to 1, 0, or $-1$ depending on whether the relative brightness of pixel $i$ is greater, the same, or lower than that of pixel $j$.

\subsection{Smoothness priors}

 The albedo component is constrained using a multi-scale $L_1$ smoothness term:
\begin{equation}
    \begin{aligned}
        \mathcal{L}_\mathrm{asmooth} &= \\
        &\sum_{i=1}^{L} \frac{1}{N_{l}l} \sum_{i=1}^{N_l}\sum_{j\in \mathcal{N}(l,i)} v_{l,i,j} ||\log A_{l,i} - \log A_{l,j}||_1,
    \end{aligned}
    \label{albedo_smooth}
\end{equation}
in which $\mathcal{N}(l,i)$ indicates the 8-connected neighborhood of the pixel at position $i$ at scale $l$. $v_{l,i,j}$ is the weight corresponding to the similarity between the pair of albedo pixels $(i,j)$, which is formulated as $\exp(-\frac{1}{2}(\mathbf{f}_{l,i} - \mathbf{f}_{l,j})^T \Sigma^{-1}(\mathbf{f}_{l,i} - \mathbf{f}_{l,j}))$. The greater the similarity of the two pixels, the smaller the weight, making the pairwise term loss less significant. $\mathbf{f}_{l,i}$ is the feature vector defined as $[\mathbf{p}_{l,i}, I_{l,i}, c_{l,i}^1, c_{l,i}^2]$, where $\mathbf{p}_{l,i}$ is spatial position, $I_{l,i}$ is  image intensity, and $c_{l,i}^1$ and $c_{l,i}^2$ are the first two elements of chromaticity. $\Sigma$ is the covariance matrix defining the distance between two feature vectors. This albedo smoothness term encourages the reconstructed albedo layer  to be piecewise constant.

The shading smoothness is formulated using a densely-connected $L_2$ term:
    \begin{equation}
        \begin{aligned}
            \mathcal{L}_\mathrm{ssmooth} = \frac{1}{2N}\sum_{i}^{N} \sum_{j}^{N} \hat{W}_{i,j}(\log S_i - \log S_j)^2,
        \end{aligned}
        \label{shading_smooth}
    \end{equation}
where $\hat{W}$ is a bi-stochastic weight matrix derived from $W$ where
\[
W_{i,j}=\exp(-\frac{1}{2}||\frac{\mathbf{p}_i - \mathbf{p}_j}{\sigma_p}||_2^2).\] 
A detailed derivation can be found in \cite{li2018-watching, barron2015fast}.
This weight is used to measure the positional difference of a pair of pixels $(i, j)$ in an image, with greater weight for nearby pixels.
%    This weight relates to the pixel position difference of the pixel pair $(i,j)$, making this term stronger in nearby area around pixel $i$.

\section{Further results}
\begin{figure*}[!tp]
    \centering
    \includegraphics[width=1.0\linewidth]{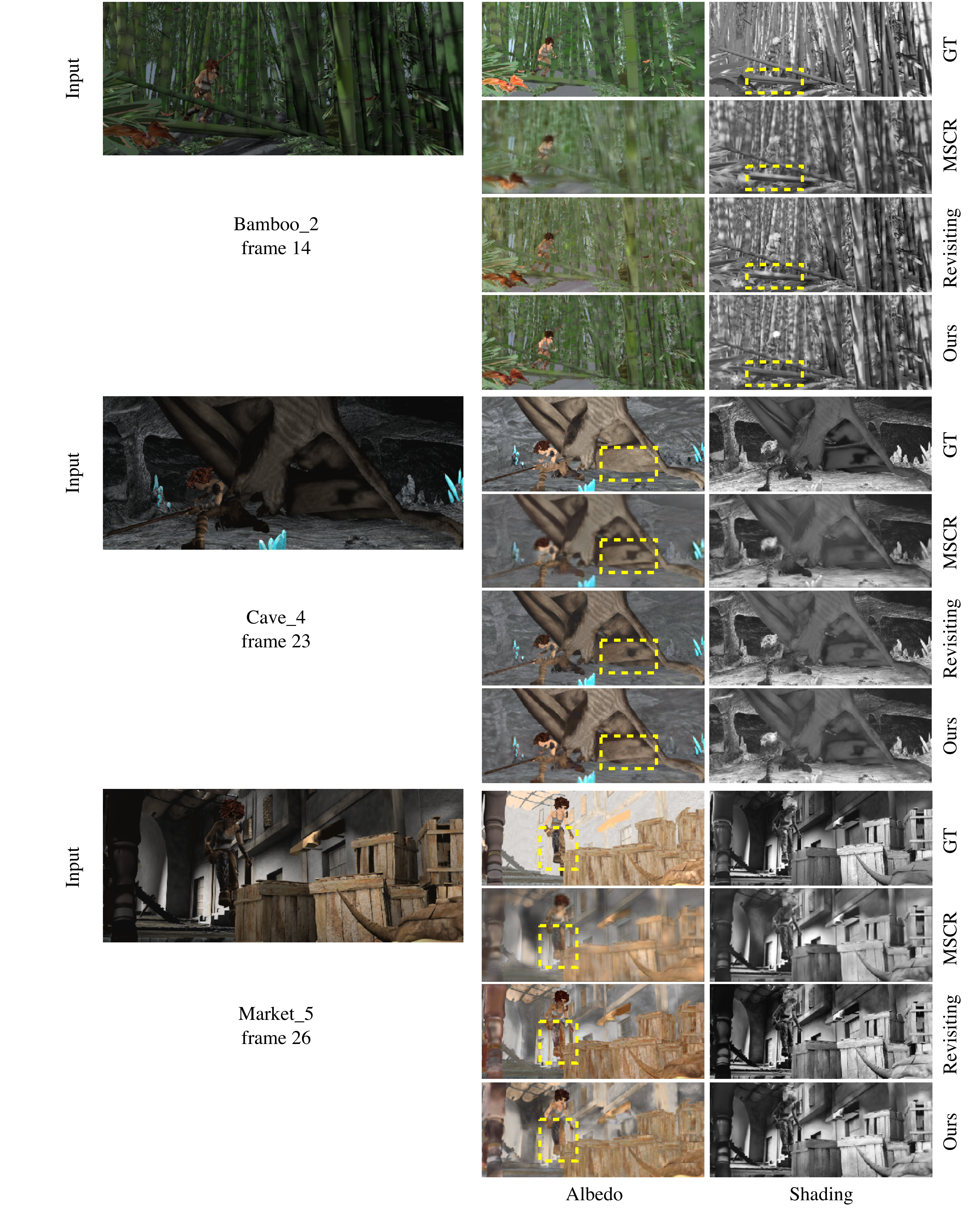}    \caption{A further qualitative comparison on our refined MPI Sintel dataset MPI\_RD, using the scene split. Our method is better at separating albedo and shading components. The regions with obvious differences are highlighted in dotted boxes.
    }
    \label{fig:12_more_MPI_results_2}
\end{figure*}

\begin{figure*}[!tp]
    \centering
    \includegraphics[width=1.0\linewidth]{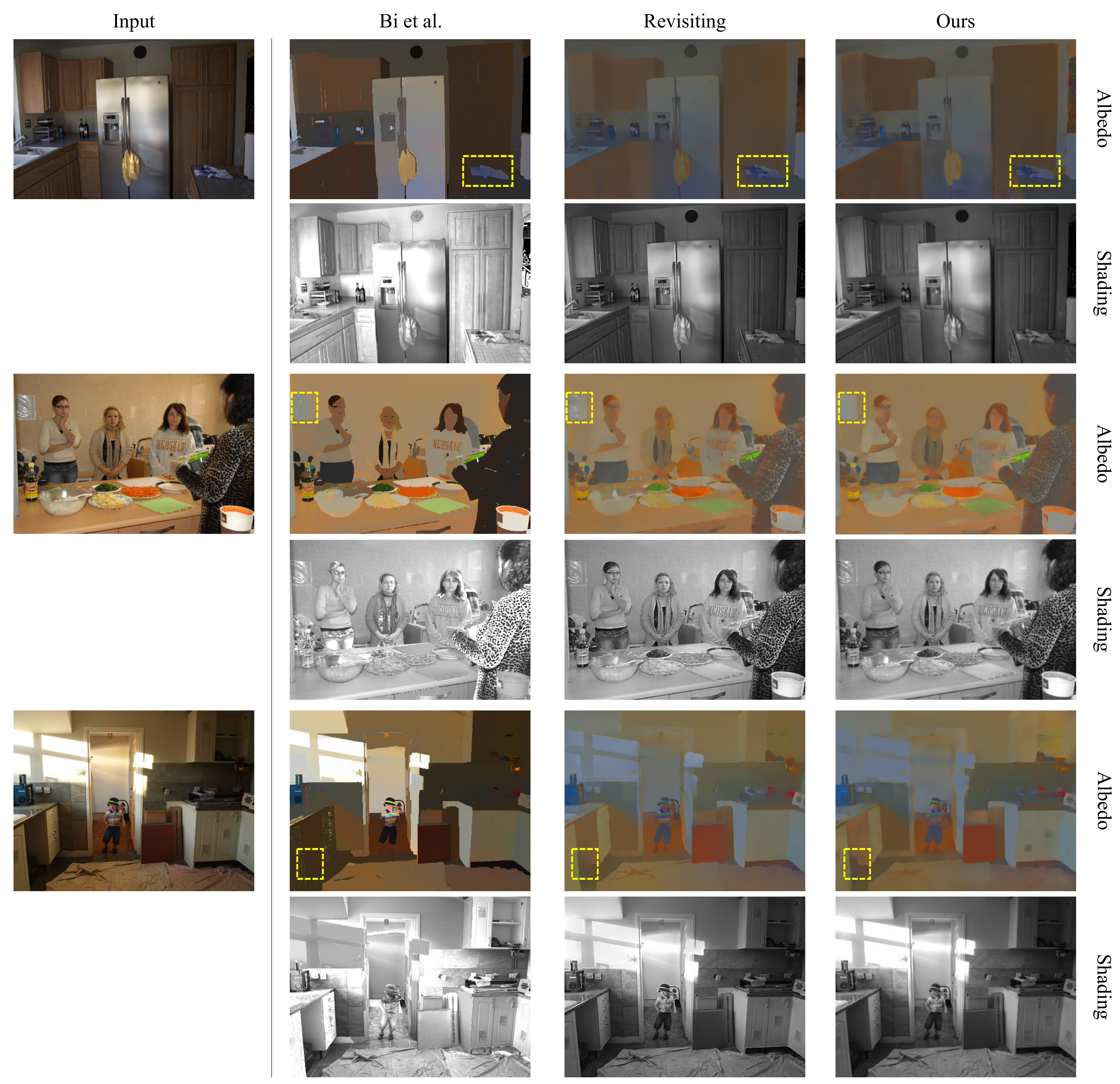}    \caption{A further qualitative comparison on the IIW dataset. 
        Regions with obvious differences are highlighted in dotted boxes.
    }
    \label{fig:13_more_IIW_results}
\end{figure*}

Further qualitative comparisons are shown in Figure~\ref{fig:12_more_MPI_results_2}, using the refined MPI Sintel dataset and Figure~\ref{fig:13_more_IIW_results}, using the IIW dataset.

%\bibliographystyle{IEEEtran}


\begin{IEEEbiography}[{\includegraphics[width=1in,height=1.25in,clip,keepaspectratio]{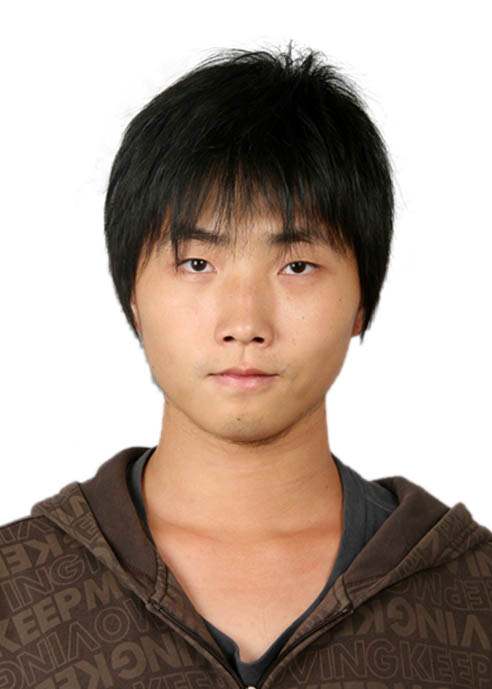}}]{Zongji Wang}
received his B.S. degree in the School of Mathematics and Systems Science at Beihang University in 2014, and his Ph.D. degree in the School of Computer Science and Engineering at Beihang University in 2021. He is currently an Assistant Professor in the Key Laboratory of Network Information Systems Technology (NIST), Aerospace Information Research Institute, Chinese Academy of Sciences. His research interests include computer vision and computer graphics.
\end{IEEEbiography}

\begin{IEEEbiography}[{\includegraphics[width=1in,height=1.25in,clip,keepaspectratio]{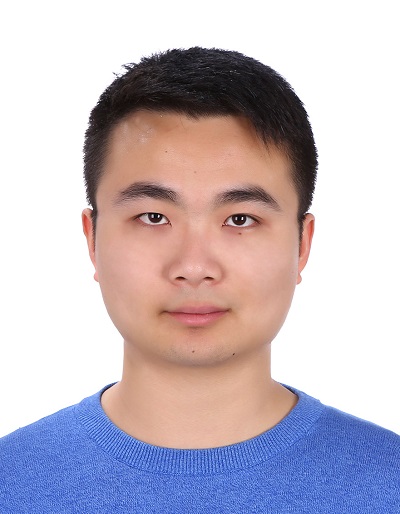}}]{Yunfei Liu}
is currently pursuing a Ph.D. degree in the State Key Laboratory of Virtual Reality Technology and Systems, School of Computer Science and Engineering, Beihang University. His research interests include computer vision, computational photography, and image processing.
\end{IEEEbiography}

\begin{IEEEbiography}[{\includegraphics[width=1in,height=1.25in,clip,keepaspectratio]{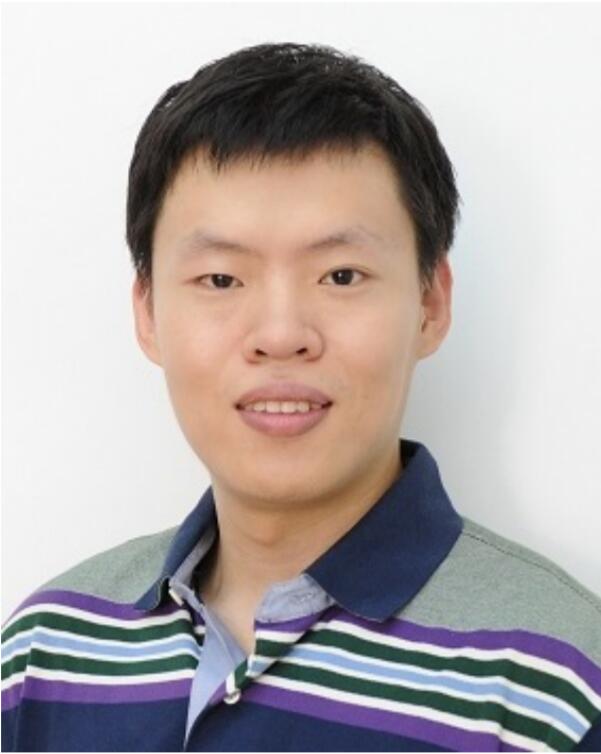}}]{Feng Lu}
received his B.S. and M.S. degrees in automation from Tsinghua University, in 2007 and 2010, respectively, and his Ph.D. degree in information science and technology from the University of Tokyo, in 2013. He is currently a Professor with the State Key Laboratory of Virtual Reality Technology and Systems, School of Computer Science and Engineering, Beihang University. His research interests include computer vision, human-computer interaction and augmented intelligence.
\end{IEEEbiography}


%\vfill


\end{document}